\documentclass[a4paper,10pt,journal]{IEEEtran}
\usepackage{epsfig}
\usepackage{graphicx}
\usepackage[cmex10]{amsmath}
\usepackage{amssymb}
\usepackage{fancybox}
\usepackage{alltt}
\usepackage{soul}
\usepackage{color}
\usepackage{verbatim}
\usepackage{xcolor}
\usepackage{colortbl,hhline}
\usepackage{algorithmic}
\usepackage[ruled,vlined]{algorithm2e}
\usepackage{framed}
\usepackage{amsthm}
\usepackage{fancyref}

\def \lb {{\langle}}
\def \rb {{\rangle}}
\newcommand{\fro}[1]{\|#1\|_2}

\usepackage[pagebackref=true,breaklinks=true,letterpaper=true,colorlinks,
bookmarks=false]{hyperref}

\newtheorem{theorem}{Theorem}

\newtheorem{proposition}[theorem]{Proposition}

\newtheorem{definition}[theorem]{Definition}

\newcommand{\grad}{\textrm{grad} }
\newcommand{\dive}{\textrm{div} }

\begin{document}

\title{Total variation regularization for fMRI-based prediction of behaviour}

\author{\IEEEauthorblockN{Vincent
Michel\IEEEauthorrefmark{1}\IEEEauthorrefmark{3}$^{1}$\thanks{$^{1}$ 
Contributed equally.
\newline
},
Alexandre Gramfort\IEEEauthorrefmark{1}$^{1}$,
Ga\"el Varoquaux\IEEEauthorrefmark{1},
Evelyn Eger\IEEEauthorrefmark{2} and
Bertrand Thirion\IEEEauthorrefmark{1}}
\\
\IEEEauthorblockA{
\IEEEauthorrefmark{1}INRIA, Saclay-\^Ile-de-France, Parietal
team, France - CEA/ DSV/ I2BM/ Neurospin/ LNAO\\}
\IEEEauthorblockA{\IEEEauthorrefmark{2}
INSERM U562, France -
CEA/ DSV/ I2BM/ Neurospin/ Unicog\\}
\IEEEauthorblockA{\IEEEauthorrefmark{3}Corresponding author -
vincent.michel@inria.fr}}

\maketitle

\begin{abstract}
While medical imaging typically provides massive amounts of
data, the extraction of relevant information for predictive diagnosis
remains a difficult challenge.
Functional \emph{MRI} (\emph{fMRI}) data, that provide an indirect measure of
task-related or spontaneous neuronal activity, are classically analyzed in a
mass-univariate procedure yielding statistical parametric maps.
This analysis framework disregards some important principles of brain
organization: population coding, distributed and overlapping
representations.
Multivariate pattern analysis, \emph{i.e.}, the prediction of behavioural
variables from brain activation patterns better captures this
structure.
To cope with the high dimensionality of the data, the
learning method has to be regularized.
However,  the spatial structure of the image is not taken into account
in standard regularization methods, so that the extracted features are often
hard to interpret.
More informative and interpretable results can be
obtained with the $\ell_1$ norm of the image gradient, \emph{a.k.a.}
its Total Variation (TV), as regularization.
We apply for the first time this method to fMRI data, and show that
TV regularization is well suited to the purpose of brain
mapping while being a powerful tool for brain decoding.
Moreover, this article presents the first use of TV regularization
for classification.

\end{abstract}


\section{Introduction}
\label{sec:intro}

Functional Magnetic Resonance Imaging (or fMRI) has been widely used
for more than fifteen years for neuroscientific and cognitive studies.
The analysis of these data largely relies on the general linear model
(GLM), introduced for functional imaging by Friston et al.
\cite{friston1995c}. The
GLM is a simple yet powerful framework for 
deciding which brain regions exhibit a significantly positive
task-related effect.
This inference, also called \emph{classical inference}, is based on
statistical tests applied to each voxel separately, yielding
significance maps (\emph{a.k.a.} Statistical Parametric Maps -
SPMs) for the effects under consideration.
However, despite its simplicity and the accuracy of the SPMs, classical
inference suffers from a major drawback: it analyzes each
voxel separately and consequently cannot fully exploit the correlations
existing between different brain regions to improve the inference. Spatial
information is only taken into account in testing procedures, \emph{e.g.} by
using the cluster size tests in Random Field Theory.

Correlations between brain activations are likely to arise as a
consequence of processing in distributed populations of
neurons~\cite{georgopoulos1986,tsao2006,nevado2004}.  
This is particularly the case in \emph{population coding}
models~\cite{pouget2000,dayan2001}.
For the purpose of statistical inference, these models suggest
that effects that differ between experimental conditions 
are not optimally characterized by the effect significance
at individual voxels~\cite{kanwisher2000},
and that one should rather consider the combined information from
different voxels/regions of the brain \cite{sidtis2003}.
Moreover, statistical power in the case of classical inference is
limited by the multiple comparison problem (one statistical test is
performed for each voxel and the number of comparisons has to be
corrected for).
\renewcommand{\labelitemi}{$\bullet$}

Recently, the inference of behavioral information or cognitive states
from brain activation images such as those obtained with fMRI has
emerged as an alternative neuroimaging data analysis
paradigm~\cite{cox2003,laconte2005,otoole2007}.  
It can be used to assess the specificity of several brain regions for
certain cognitive or perceptual functions, by evaluating the accuracy
of the prediction of a behavioral variable of interest -- the
\textit{target} -- based on the activations measured in these regions.
This inference relies on a prediction function, the accuracy of which
depends on whether it uses the relevant variables, \emph{i.e.,} the
correct brain regions.
This approach, called \emph{inverse inference}, has some major advantages:
\begin{itemize}
\item As multivariate approach, it is consistent with 
population coding models. Indeed, the neural information, which can be
encoded by different populations of neurons, can be decoded using a
\emph{pattern} of voxels \cite{cox2003,haynes2006}.
\item It avoids the multiple comparison issue, as it performs only one
statistical test (on the predicted behavioral variable). In that sense, it can
detect significant links between image data and target that would not have been
detected by standard statistical parametric mapping
procedures~\cite{kamitani2005}; note however that the statistical
interpretation of these two tests are clearly different.
\item It addresses new challenges, in particular by allowing to identify a new
stimulus in a large dataset, based on already seen stimuli (as visual stimuli
\cite{kay2008}, or nouns associated with new images \cite{mitchell2008}).
Moreover, it can be used for the more challenging generalization of the
prediction to unknown high level stimuli~\cite{knops2009}, which opens
a deeper understanding on brain functional organization.
\end{itemize}

Many machine learning methods have been applied to
fMRI activation images. Among them are linear
discriminant analysis~\cite{cox2003},
support/relevance vector machines \cite{laconte2005},
neural networks \cite{haxby2001},
\emph{Lasso} \cite{liu2009}, elastic net
regression \cite{carroll2009}, kernel ridge regression \cite{chu2010},
boosting \cite{martinezramon2006}, sparse logistic regression
\cite{ryali2010,raizada2010} or Bayesian regularization
\cite{yamashita2008,friston2008,michel2009}. Moreover, fMRI data are
intrinsically smooth, so that their spatial structure has to be taken into
account. Spatial
information has thus been considered within the inverse inference
framework, by using specific priors
in a Bayesian framework \cite{palatucci2007} or by creating spatially
informed features \cite{michel2010}.
In the inverse inference problem the main objective remains the
extraction of informative regions within the brain volume (see
\cite{haynes2006} for a review). 
Besides prediction accuracy, an even greater challenge in brain functional
imaging, is the ability of the method to provide an
interpretable model (see e.g. \cite{carroll2009}). Ultimately, the
predictive function learned from the data should be as explicit as
standard statistical mapping results.  This double objective is
addressed by the present contribution.

In practice, selecting the relevant voxels -- called features in
machine learning -- is fundamental in order to achieve accurate
prediction. However, when the number of \textit{features} (voxels) is
much larger than the numbers of \textit{samples} (images), the
prediction method may overfit the training set. In other words, it
fits seemingly predictive information from noise in the training set,
and thus does not generalize well to new data.
To address this issue, one can reduce the number of
features.  A classical strategy consists of preceding the learning
algorithm with a feature selection procedure that drastically reduces
the spatial support of predictive regions.  To date, the most widely
used method for feature selection is voxel-based \emph{Anova}
(Analysis of Variance), that evaluates each voxel independently. This
is often combined with the use of Support Vector Machine as prediction
function (see
\cite{mitchell2004,laconte2005,mouraomiranda2005,wang2007,hanson2008}).
An alternative approach consists in performing the model estimation by
taking the high dimensional data as input while using relevant regularization
methods. These regularizations are
performed with two possible goals: stabilizing the estimation of the weights of
the features, and/or forcing a majority of features to have close to zero
weights (\emph{i.e.} promoting sparsity).

Let us introduce the following predictive linear model:
\begin{equation}
\bold{y} = f(\bold{X},\bold{w}, b) = F(\bold{X w} + b) \enspace ,
\label{Eq:predictive_model}
\end{equation}
where $\bold{y}$ represents the behavioral variable and $(\bold{w},b)$ are the
parameters to be estimated on a training set.
A vector $\bold{w} \in \mathbb{R}^p$ can be seen as an image;
$p$ is the number of features (or voxels) and $b \in \mathbb{R}$
is called the \emph{intercept}. The matrix $\bold{X} \in \mathbb{R}^{n\times p}$
is the design matrix. Each row is a $p$-dimensional sample, \emph{i.e.},
an activation map related to the observation.
It has been shown \cite{cox2003,laconte2005} that using a non-linear
classifier does not improve the prediction accuracy, and yields interpretation
issues. Thus, we only focus on linear classifiers in
this paper.
Depending on whether the variable to be predicted takes scalar or discrete
values, the learning problem is either a regression or a classification
problem. In a linear regression setting, $f$ reads:
\begin{equation}
f(\bold{X},\bold{w},b) = \bold{X}\, \bold{w} + b \enspace ,
\label{Eq:model_reg}
\end{equation}
with $\bold{y} \in \mathbb{R}^n$.
In the case of classification with a linear model, $f$ is defined by:
\begin{equation}
f(\bold{X},\bold{w},b) = \mbox{sign} (\bold{X}\, \bold{w} + b) \enspace ,
\label{Eq:model_classif}
\end{equation}
where ``$\mbox{sign}$'' denotes the sign function and $\bold{y} \in
\{-1,1\}^n$.

The crucial issue here is that $n \ll p$, so that estimating $\bold{w}$ is an
ill-posed problem. The estimation requires therefore adapted
regularization.
A standard approach to perform the estimation of $\bold{w}$ with
regularization uses penalization of a maximum likelihood estimator.
It leads to the following minimization problem:
\begin{equation}
\bold{\hat{w}} = \mathrm{arg}\!\min_{\bold{w},b}\;
\mathcal{L}(\bold{y},F(\bold{X}
\bold{w}+b)) + \lambda J(\bold{w}) \;\;,\;\; \lambda \geq 0
\label{eq:opt_pb}
\end{equation}
where $\lambda J(\bold{w})$ is the regularization term and
$\mathcal{L}(\bold{y},F(\bold{X} \bold{w}+b))$ is
the loss function. The parameter $\lambda$
balances the loss function and the penalty $J(\bold{w})$. Note that
the intercept $b$ is not included in the regularization term.

The use of the intercept is fundamental in practice as it allows the
separating hyperplane to be offset from 0. 
However for the sake of simplicity in the presentation of the method,
we will from now on consider $b$ as an added coefficient in the vector
$\bold{w}$. This is classically done by concatenation of a column
filled with 1 in the matrix $\bold{X}$. The loss function will also be
abbreviated $\mathcal{L}(\bold{w})$.

In the formalism of \eqref{eq:opt_pb}, the reference method is \textit{elastic
net}
\cite{hastie2005},
which is a combined $\ell_{1}$ and $\ell_{2}$ penalization:
\begin{equation}
\lambda J(\bold{w}) = \lambda_1 \|\bold{w}\|_{1}  + \lambda_2
\|\bold{w}\|^{2}_{2} = \sum_{i = 1}^{p} \lambda_1 |w_i| + \lambda_2
w_i^2
\label{eq:enet}
\end{equation}
\textit{Elastic net} has two limit cases:
$\lambda_2 = 0$ is the \textit{Lasso} \cite{tibshirani1996} which
yields an extreme sparsity in the selected features,
and $\lambda_1 = 0$ corresponds to \textit{Ridge regression}
\cite{kennard1970}.

A major limitation of the methods cited above, including the
latter penalization, is that they do not take
into account the underlying
structure of $\bold{w}$. In the
case of brain images, $\bold{w}$ is defined on
a spatial 3-dimensional grid.
The main motivation for using this spatial structure is that the
predictive information is  most likely organized in regions, and not randomly
spread across voxels~\cite{thirion2006,michel2010}.
As it is demonstrated in this contribution, one can both decrease the complexity
of the results (\emph{i.e} increase the
\emph{interpretability} of the results by extracting a small set of spatially
coherent regions of interest) as well as increase the \emph{accuracy} of the
prediction by taking into account the spatial relations between voxels.

In this article, we develop an approach for regularized prediction
based on Total Variation (TV), $J(\bold{w}) = TV(\bold{w})$.
TV, mathematically defined as the $\ell_1$ norm of
the image gradient, has been primarily used for image denoising
\cite{rudin1992,chambolle2004} as it preserves edges. The motivation for using
TV for brain imaging is that it promotes
estimates $\bold{\hat{w}}$ of $\bold{w}$
with a block structure, creating regions with piecewise constant weights,
and therefore outlining the brain regions correlated to the target behavioral
variable. Indeed, we are expecting that
the spatial layout of the neural code is sparse and spatially structured in
the
sense that non-zero weights are grouped into connected clusters. Weighted maps
showing such characteristics will be called interpretable, as they fulfill our
hypothesis on the spatial layout of neural coding \cite{chklovskii2004}.
This approach is closely related to the one developed in
\cite{stoeckel2005}, that introduce proximity  information about the
features in the regularization term.

In this contribution, the mathematical and implementation details of
\emph{TV regression/classification} are first detailed. As far as we know, the
present work is the first to use TV in the context of image
classification and also the first one to propose the use of the image structure
in the learning framework of \eqref{eq:opt_pb} in the context of fMRI
inverse inference.
We apply both TV regression and TV classification to an fMRI
paradigm that studied the processing of object shape and size in the human
brain.
Results show that TV outperforms other state of the art methods, as it yields
better prediction performance while providing weights $\bold{\hat{w}}$ with an
interpretable spatial structure.

\section{Total Variation and prediction}

We first detail the notations of the problem. We then develop
the TV regularization framework. Finally, we detail the algorithm
used for regression and classification.

\subsection{Notations}

Let us define $\Omega \subset \mathbb{R}^3$ the 3D image domain, discretized on
a finite grid. The coefficients $\bold{w}$ define a function
from $\Omega$ to $\mathbb{R}$, \emph{i.e.}, $\bold{w}: \Omega \rightarrow
\mathbb{R}$. Its TV reads:
\begin{eqnarray*}
TV(\bold{w}) &=& \sum_{\omega \in \Omega} \|\nabla \bold{w}\|(\omega) \\
&=& \sum_{\omega \in
\Omega}
\sqrt{\nabla_{x} \bold{w}(\omega)^2 + \nabla_{y} \bold{w}(\omega)^2 +
\nabla_{z}\bold{w}(\omega)^2}
\end{eqnarray*}
Let us assume that $\omega$ stands for the voxel at position $(i,j,k)$,
away from the border of $\Omega$,
then $\nabla_{x} \bold{w}(\omega)^2$ corresponds to
$(\bold{w}_{i+1,j,k} - \bold{w}_{i,j,k})^2$ (see appendix
\ref{appendix_grad} for more details).
TV can be used with different discretizations, such as an anisotropic
discretization. However, such a discretization is biased in the direction
of the axes of the image, which is problematic especially with a strong
regularization. Indeed, an isotropic discretization promotes sparse gradient
along the image axes. We use therefore the standard isotropic discretization
of TV~\cite{chambolle2004,goldfarb2009}.

We denote $\bold{y} \in \mathbb{R}^{n}$ the targets to be predicted, and
$\bold{X}\in\mathbb{R}^{n \times p}$ the set of activation images related to the
presentation of different stimuli. The integer $p$ is the number of voxels and
 $n$ the
number of samples (images). Typically, $p \sim 10^3 \text{ to } 10^5$ (for a
whole volume), while $n \sim 10 \text{ to } 10^2$.
We denote $M$ the mask of the brain that comes from standard fMRI analysis,
and that is used to avoid computation outside of the brain volume. $M$ is
a $p_i \times p_j \times p_k$ three dimensional grid, with:
\begin{eqnarray*}
\begin{cases}
    M_{i,j,k} = 1 & \mbox{if the voxel is in the mask}\\
    M_{i,j,k} = 0 & \mbox{if the voxel is not in the mask}
\end{cases}
\end{eqnarray*}
with $\sum_{i,j,k} M_{i,j,k} = p$.
Additionally, we define  $\grad: \mathbb{R}(\Omega) \rightarrow
\mathbb{R}^3(\Omega)$ a gradient operator and
 $\dive: \mathbb{R}^3(\Omega)
\rightarrow \mathbb{R}(\Omega)$
the associated adjoint divergence operator (the adjoint
operator is used in the convex optimization algorithm, see appendix
\ref{appendix_grad} for more details, in particular
Eq.~\ref{Eq:adjoint}).

 Let $K$ the convex
set defined
by:
\begin{eqnarray*}
K &=& \{ g : \Omega \rightarrow \mathbb{R}^3 \; |\; \forall \omega \in \Omega,
\|g(\omega)\| \; \leq 1 \}
\end{eqnarray*}
and $\Pi_{K}$ the projection operator onto the set $K$:
\begin{eqnarray*}
\begin{cases}
    \Pi_{K}(g)(\omega) = g(\omega) & \mbox{if\ } \|g(\omega)\| \leq 1\\
    \Pi_{K}(g)(\omega) = g(\omega)/\|g(\omega)\| & \mbox{otherwise.}
\end{cases}
\end{eqnarray*}
This projection operator will be used
in the optimization loop solving Eq.~\ref{prop:dual_prox_tv}, to apply the
constraint. It can be viewed as the projection on the $\ell_\infty$ norm (dual
of the $\ell_1$ norm) ball.

\subsection{Convex optimization}

We consider the minimization problem \eqref{eq:opt_pb}.
When $J(\bold{w})$ is non-smooth (\emph{i.e.}
not differentiable), an analytical solution does not exist
and the optimization can unfortunately not
be performed with simple algorithms such as Gradient descent
and Newton method. This is for example the case with
$J(\bold{w})=\|\bold{w}\|_1$ ($\ell_1$ norm \emph{a.k.a.} \emph{Lasso} penalty)
and with $J(\bold{w})=TV(\bold{w})$,
both of which require advanced optimization strategies.

A recently studied strategy
(\cite{daubechies2004,combettes2005,nesterov2007,beck2009a}) is
based on iterative procedures involving the computation of
\emph{proximity operators} (see def. \ref{Def:proximity})
\cite{moreau1965}. Such approaches are adapted to composite problems
with both a smooth term and a non-smooth term as it is the case here
(see \cite{tseng2009} for a recent review).
In the context of neuroimaging, such optimization schemes have been proposed
recently in order to solve the inverse problem of magneto- and
electro-encephalography (collectively M/EEG) when considering non $\ell_2$
priors \cite{gramfort2009a,gramfort2009b}.

\begin{definition}[Proximity operator]
Let $J:\mathbb{R}^p\rightarrow \mathbb{R}$ be a proper convex
function. The proximity operator associated  with $J$ and $\lambda \in
\mathbb{R}_{+}$ denoted by $\mbox{prox}_{\lambda J} :
\mathbb{R}^{p}\rightarrow\mathbb{R}^p$ is given by:
$$
\mbox{prox}_{\lambda J}(\bold{w}) = \mathrm{arg}\min_{\bold{v}
\in\mathbb{R}^p}
\left(\frac{1}{2}\|\bold{v} -\bold{w}\|_2^2 + \lambda J(\bold{v}) \right)
$$
\label{Def:proximity}
\end{definition}

The iterative procedure known as \emph{ISTA} (\emph{Iterative
Shrinkage-Thresholding Algorithm}, \emph{a.k.a}
\emph{Forward-Backward iterations}) \cite{daubechies2004,combettes2005},
is based on the alternate minimization of the \emph{loss} term
$\mathcal{L}(\bold{w})$, by gradient descent, and the penalty $J(\bold{w})$,
by computing a proximity operator.
One can show (see appendix \ref{appendix_prox} for a sketch of the proof),
that this can be done in one single step by iterating:
\begin{equation}
\bold{w}^{(k+1)} =  \mbox{prox}_{\lambda J /L} \left(\bold{w}^{(k)}
- \frac{1}{L}\nabla  \mathcal{L}(\bold{w}^{(k)}) \right) \;\; ,
\label{Eq:prox_ista}
\end{equation}
where $\frac{1}{L}\nabla  \mathcal{L}(\bold{w}^{(k)})$ is the gradient descent
term
with a
stepsize $\frac{1}{L}$, $\mbox{prox}_{\lambda J /L}$ is the proximity operator
of the penalty and
the scalar $L$ is an upper bound on the \emph{Lipschitz constant}
$L_0$ of the gradient of the loss function.
The pseudo code of the \emph{ISTA} procedure is defined in
Algo.~\ref{Tab:pseudocode_ista}.

Inspired by previous findings~\cite{nesterov2007},
the \emph{FISTA} (\emph{Fast Iterative Shrinkage-Thresholding
Algorithm}) procedure \cite{beck2009a,beck2009b} has been
developed to speed up the convergence of \emph{ISTA}. While \emph{ISTA}
converges in $\mathcal{O}(1/K)$, \emph{FISTA} is proven to converge in
$\mathcal{O}(1/K^2)$, where $K$ is the number of iterations.
The pseudo code of the \emph{FISTA} procedure is given in
Algo.~\ref{Tab:pseudocode_fista}.
The main improvement in \emph{FISTA} is to compute the next descent direction
using the previous one. Such an idea is also present in the well
known conjugate gradient algorithm that uses all previous iterates
to compute the next descent direction.

\begin{algorithm}[htb]
\caption{\emph{ISTA} procedure}
Compute the Lipschitz constant $L_0$ of
the operator $\nabla  \mathcal{L}$.\\
Initialize $\bold{w}^{(0)} \in \mathbb{R}^p$\\
\Repeat{convergence}{
    $
    \bold{w}^{(k+1)} =  \mbox{prox}_{\lambda J /L} \left(\bold{w}^{(k)}
                            - \frac{1}{L}\nabla  \mathcal{L}(\bold{w}^{(k)})
\right)
    $\\
    where $L > L_0$.
}
\Return $\bold{w}$
\label{Tab:pseudocode_ista}
\end{algorithm}
\begin{algorithm}[htb]
\caption{\emph{FISTA} procedure}
Compute the Lipschitz constant $L_0$ of
the operator $\nabla  \mathcal{L}$.\\
Initialize $\bold{w}^{(0)}\in \mathbb{R}^p$, $\bold{v}^{(1)}=\bold{w}^{(0)}$ and
$t_1 = 1$.\\
\Repeat{convergence}{
\begin{eqnarray*}
\bold{w}^{(k)} &=&  \mbox{prox}_{\lambda J /L} \left(\bold{v}^{(k)}
                        - \frac{1}{L}\nabla \mathcal{L}(\bold{v}^{(k)})
\right)\\
t_{k+1} &=& \frac{1+\sqrt{1+4t_k^2}}{2}\\
\bold{v}^{(k+1)} &=& \bold{w}^{(k)} + \left( \frac{t_k - 1}{t_{k+1}}
\right)(\bold{w}^{(k)}-\bold{w}^{(k-1)})
\end{eqnarray*}
}
\Return $\bold{w}$
\label{Tab:pseudocode_fista}
\end{algorithm}

Let us introduce now the notion of \emph{Duality gap}.
The duality gap is a natural stopping condition for approaches
as \emph{ISTA} and \emph{FISTA}.
In practice, if the duality gap is below a
value $\epsilon>0$, it guarantees that the solution obtained is
$\epsilon$-optimal, \emph{i.e.,}
that the value of the cost-function reached by the algorithm
is not greater than $\epsilon$ more the globally optimal value.
A comprehensive presentation of this notion~\cite{boyd2004} is beyond the
scope of this paper, and we now give some details in the particular
case of the proximity operator $\mbox{prox}_{\lambda TV}$
known as the \emph{ROF} problem \cite{rudin1992} (named after the authors
L. Rudin, S. Osher and E. Fatemi) in the image processing literature.

The computation of $\mbox{prox}_{\lambda TV}$ and the associated duality gap
requires the derivation of a Lagrange dual problem~\cite{boyd2004}.

\begin{proposition}[$\mbox{prox}_{\lambda TV}$ Dual problem]
A dual problem associated with $\mbox{prox}_{\lambda TV}$ is given by
\begin{equation}
    \bold{z}^* = \mathrm{arg}\!\max_{\bold{z} \in K} - \| \dive \, \bold{z} +
\bold{w} / \lambda \|_2^2 \enspace ,
\label{eq:dual_prox_tv}
\end{equation}
where $z$ is the dual variable that satisfies $\bold{v}^* = \bold{w} +
\lambda
\, \dive \, \bold{z}^*$, with $\bold{v}^* = \mbox{prox}_{\lambda
TV}(\bold{w})$
\label{prop:dual_prox_tv}
\end{proposition}

This result is adapted from \cite{chambolle2004} (see appendix
\ref{appendix_dualgap} for a sketch of the proof). The problem
\eqref{eq:dual_prox_tv} is a maximization of a smooth concave function
over a convex set. As shown in \cite{beck2009b}, it can be solved with
the \emph{FISTA} iterative procedure.  The resolution of the \emph{ROF}
problem is therefore achieved by solving the dual problem. Once
$\bold{z}^*$ is obtained, $\bold{v}^* = \mbox{prox}_{\lambda
TV}(\bold{w})$ is given by $\bold{v}^* = \bold{w} + \lambda \, \dive
\, \bold{z}^*$.

The latter result also gives an estimate of the duality gap.

\begin{proposition}[Duality gap]
The duality gap $\delta_{gap}$ associated with the \emph{ROF} problem is
given by:
\begin{equation}
    \delta_{gap}(\bold{v}) = \frac{1}{2}\fro{\bold{w}-\bold{v}}^2 + \lambda TV
(\bold{v})
    - \frac{1}{2}(\fro{\bold{w}}^2 + \fro{\bold{v}}^2) \geq 0 \enspace ,
\label{Eq:dualgap_def}
\end{equation}
where the primal variable $\bold{v}$ is obtained during the iterative
procedure from the current estimate of the dual variable $\bold{z}$
with $\bold{v} = \bold{w} +
\lambda \, \dive \, \bold{z}$.
\label{Prop:dualgap}
\end{proposition}
\noindent
See appendix \ref{appendix_dualgap} for more details.
This duality gap will be used as a stopping criterion for the
\emph{FISTA} procedure solving the \emph{ROF} problem. At each
iteration of the \emph{FISTA} procedure, we will stop the iterative
loop if the duality gap is below a given threshold
$\epsilon$. 
In practice, $\epsilon$ is set to $10^{-4} \times \|
\bold{w} \|_2^2$ to be invariant to the scaling of the data.

Note that the \emph{ROF} problem can be also solved using very efficient
combinatorial optimization methods \cite{goldfarb2009}, when using the
anisotropic discretization of TV.

\subsection{Prediction framework}

We now detail the original contribution of this work, that is the
construction of a predictive framework using the TV regularization.
For $J(\bold{w})= TV(\bold{w})$, the global algorithm for solving the
minimization problem defined in \eqref{eq:opt_pb} consists in a
\emph{FISTA} procedure (resolution of the \emph{ROF} problem) nested
inside an \emph{ISTA} procedure (resolution of the main minimization
problem). 
The \emph{FISTA} procedure is performed at each step of
\emph{ISTA} with a \emph{warm restart} on the dual variable
$\bold{z}$.
We do not use \emph{FISTA} for solving the main minimization problem,
as this procedure requires an exact proximity operator.
The resolution of the \emph{ROF} problem only leads to an
$\epsilon$-optimal solution.
The pseudo-code of the global algorithm for the TV regularization is
provided in Table \ref{Tab:PseudoCode_reg}.

A difficulty specific to fMRI data is the computation of the gradient
and divergence over a mask of the brain with correct border conditions
(see appendix \ref{appendix_grad} for details).
Moreover, with such an irregular domain, the upper bound $\tilde{L}$
for the Lipschitz constant of the \emph{FISTA} procedure also needs to
be estimated on each input data.  To do this we use a power method
that is classically used to estimate the spectral norm of a linear
operator, here equal to the Laplacian $\Delta: \Omega \rightarrow
\Omega$ defined by $\Delta(\omega) = \dive (\grad (\omega))$.

\begin{algorithm}[h!tb]
\caption{TV regularization solver}
Set maximum number of iterations $K$ (\emph{ISTA}).\\
Set the threshold $\epsilon$ on the dual gap (\emph{FISTA}).\\
Set $L = 1.1L_0$ where $L_0$ is the Lipschitz constant of
$\nabla \mathcal{L}$.\\
Set $\tilde{L} = 1.1\tilde{L_0}$ where $\tilde{L_0}$ is the
Lipschitz constant of the Laplacian operator
$\Delta: w \in \mathbb{R}(\Omega) \rightarrow \dive(\grad(w))$.\\
Initialize $\bold{z} \in \mathbb{R}(\Omega^3)$ with zeros.\\
\textbf{\#\#\# ISTA loop \#\#\#}\\
\For{$k = 1 \dots K$}{
    $\bold{u} = \bold{w} - \frac{1}{L} \nabla \mathcal{L}(\bold{w})$\\
    \textbf{\#\#\# FISTA loop \#\#\#}\\
    Initialize $\bold{z}_{aux} = \bold{z}$, $t = 1$\\
    \Repeat{$\delta_{gap}(\bold{u} + \lambda \dive(\bold{z})) \leq \epsilon$}{
        $\bold{z}_{old} = \bold{z}$\\
        $\bold{z} =  \Pi_K \left(\bold{z}_{aux} - \frac{1}{\lambda \tilde{L}}
\grad( L \bold{u}
            + \lambda\dive(\bold{z}_{aux})) \right)$\\
        $t_{old} = t$\\
        $t = (t + \sqrt{1 + 4t^2}) / 2$\\
        $\bold{z}_{aux} = \bold{z} + \frac{t_{old}-1}{t} (\bold{z} -
\bold{z}_{old})$\\
    }
    $\bold{w} = \bold{u} + \lambda \dive(\bold{z})$
}
\Return $\bold{w}$
\label{Tab:PseudoCode_reg}
\end{algorithm}

\noindent\textbf{\emph{TV regression:}} 
The regression version of the TV is called \emph{TV
regression}.  In this case, we use the least-squares \emph{loss}:
\begin{eqnarray*}
\begin{cases}
\mathcal{L}(\bold{w}) = \frac{1}{2n}\|\bold{y} -\bold{X}\bold{w} \|^2 \\
\nabla \mathcal{L}(\bold{w})  =  - \frac{1}{n} \bold{X}^T(\bold{y} - \bold{X}
\bold{w})
\end{cases}
\end{eqnarray*}
The Lipschitz constant $L_0$ of the
operator $ \nabla \mathcal{L}$ is $L_0 = \| |\bold{X}^T\bold{X} | \|  /
n$, where $\| | . | \| $ stands for the spectral norm equal to largest
singular value.
The constant $L$ is set in practice to $L = 1.1 L_0$.

\noindent\textbf{\emph{TV classification:}} 
The classification version of the TV is called \emph{TV
classification}.  This algorithm is based on a logistic loss \cite{hastie2003}.
We now give the mathematical formulation for the binary case
with $\bold{y} \in \{-1,1\}^n$. The logistic regression model defines the
conditional
probability of $y_i$ given the data $\bold{x}_i$ as:
\begin{equation}
p(y_i|\bold{x}_i,\bold{w}) = \frac{1}{1 + \exp^{-y_i (\bold{x}_i^T
\bold{w})}}
\label{Eq:log_reg}
\end{equation}
The corresponding loss and the loss gradient read:
\begin{eqnarray*}
\begin{cases}
\mathcal{L}(\bold{w}) = \frac{1}{n} \sum_{i=1}^n \log \left( 1+\exp^{-y_i
(\bold{x_i}^T \bold{w})}
\right)\\
\nabla \mathcal{L}(\bold{w})  =  - \frac{1}{n} \sum_{i=1}^n 
\frac{y_i \bold{x_i}}{1+ \exp^{y_i (\bold{x_i}^T \bold{w})}}
\end{cases}
\end{eqnarray*}

The Lipschitz constant $L_0$ of the
operator $ \nabla \mathcal{L}$ is $L_0 = \| \bold{X} \|^2 /(4n)$.
The classification framework developed in this paper
treats the binary case with a logistic model, \emph{a.k.a.,} binomial model.
In our analysis, we expand this framework to multiclass classification using a
one-versus-one voting heuristic. The number of classifiers used is
$(k) \times (k-1) / 2$, where $k$ is the number of classes.
The predicted class is then selected as the class which yields the highest
probability across the predictions of all of the classifiers, as defined in
\eqref{Eq:log_reg}. Note that a multinomial approach could also be
used~\cite{krishnapuram2007}.
However the resulting weights $\bold{w}$ become impossible to
interpret, so that the multinomial model may not adapted to the
applicative context. Indeed, with three classes, one gets two hyperplanes
from which it is hard to draw any neuroscientific conclusions. The weights of
each binary classifier have a simpler meaning. This one-versus-one voting
heuristic is the one used in LibSVM \cite{libsvm}.

\subsection{Performance evaluation}

Our method is evaluated with a cross-validation procedure that splits
the available data into training and validation sets.
In the following, $(\bold{X}^{l},\bold{y}^{l})$
are a learning set, $(\bold{X}^{t},\bold{y}^{t})$ a test set and
$\bold{\hat{y}}^{t}=F(\bold{X}^t\bold{\hat{w}})$ refers to the predicted target,
where $\bold{\hat{w}}$
is estimated from the training set.

For regression analysis, the performance of the different models is
evaluated using $\zeta$, the ratio of explained variance:
\begin{equation*}
\zeta(\bold{y}^{t},\bold{\hat{y}}^{t}) = \frac{\mbox{var}(\bold{y}^{t}) -
\mbox{var}\left(\bold{y}^{t} -
\bold{\hat{y}}^{t}\right)}{\mbox{var}(\bold{y}^{t})}
\label{Eq:ev}
\end{equation*}
This is the amount of variability in the response that can be
explained by the model (perfect prediction yields $\zeta = 1$, while
$\zeta < 0$ if prediction is worse than chance).

For classification analysis, the performance of the different models is
evaluated using the classification score denoted $\kappa$ , classically defined
as:
\begin{equation*}
\kappa(\bold{y}^{t},\bold{\hat{y}}^{t}) =
\frac{\sum_{i=1}^{n^t}\delta(y^{t}_i,\hat{y}_i^{t})}{n^t}
\end{equation*}
where $n^t$ is the number of samples in the test set, and $\delta$ is
Kronecker's delta.

The p-values are computed using a Wilcoxon signed-rank test on the
prediction score.

\subsection{Competing methods}
In our experiments, TV regression is compared to different state of the
art regularization methods:
\begin{itemize}
\item \emph{Elastic net} regression \cite{hastie2005}, that requires
setting two parameters $\lambda_1$  and $\lambda_2$ \eqref{eq:enet}.
In our analyzes, a cross-validation procedure within the training set is
used to optimize these parameters. Here, we use $\lambda_1 \in \{0.2
\tilde{\lambda},0.1 \tilde{\lambda},0.05
\tilde{\lambda}, 0.01
  \tilde{\lambda}\}$, where $\tilde{\lambda} = \|\bold{X}^T \bold{y}\|_\infty$,
and
  $\lambda_2  \in \{0.1,0.5,1.,10.,100.\}$ ($\lambda_1$ and
$\lambda_2$ parametrize two different types of norm).
\item \emph{Support Vector Regression} (\emph{SVR}) with a linear
  kernel \cite{vapnik1995}, which is the reference method in
  neuroimaging. The $C$
  parameter is optimized by cross-validation in the range $10^{-3}$ to
  $10^{1}$ in multiplicative steps of $10$.
\end{itemize}
TV classification is compared to different state of the art
classification methods:
\begin{itemize}
\item \emph{Sparse multinomial logistic regression} (\emph{SMLR})
  classification \cite{krishnapuram2007}, that requires a double
  optimization, for the two parameters $\lambda_1$ and $\lambda_2$. A
  cross-validation procedure within the training set is used to
  optimize these parameters. Here, we use $\lambda_1 \in \{0.2
\tilde{\lambda},0.1 \tilde{\lambda},0.05
\tilde{\lambda}, 0.01
  \tilde{\lambda}\}$, where $\tilde{\lambda} =  \|\bold{X}^T \bold{y}\|_\infty$,
and
  $\lambda_2  \in \{0.1,0.5,1.,10.,100.\}$.
\item \emph{Support Vector Classification} (\emph{SVC}) with a linear
  kernel \cite{vapnik1995}, which is the reference method in
  neuroimaging. The $C$
  parameter is optimized by cross-validation in the range $10^{-3}$ to
  $10^{1}$ in multiplicative steps of $10$.
\end{itemize}
All these methods are used after an \emph{Anova}-based
feature selection as this maximizes their performance. Indeed, irrelevant
features and redundant information can decrease the accuracy of a predictor
\cite{hughes1968}.
The optimal number of voxels is selected
within the range $\{50,100,250,500\}$, through a nested cross-validation
within the training set. We do not select directly a threshold on p-value or
cluster size, but rather a number of features. Additionally, we check
that increasing the range
of voxels (\emph{i.e.} adding 2000 in the range of number of selected voxels)
does not increase the prediction accuracy on our datasets.
The parameter estimation of the learning function is also performed using a
nested
cross-validation within the training set, and thus, the cross-validation
framework is used rigorously in all the experiments of this paper.
All methods are developed in \emph{C} and used in \emph{Python}.  The
implementation of \emph{Elastic net} is based on \emph{coordinate
descent} \cite{friedman2009}, while \emph{SVR} and \emph{SVC} are based on 
LibSVM \cite{libsvm}. Methods are used from \emph{Python} via the
\emph{Scikit-learn} open source package~\cite{scikit}.

\section{Experiments}
\label{sec:experiments}

\subsection{Details on simulated data}

The simulated data set $\bold{X}$ consists of $n = 100$ images (size
$12\times12\times12$ voxels) with a set of four square Regions of
Interest (ROIs) (size $2\times2\times2$). We call $\mathcal{R}$ the
support of the ROIs (\emph{i.e.}  the $32$ resulting voxels of
interest). Each of the four ROIs has a fixed weight in
$\{-0.5,0.5,-0.5,0.5\}$. We call $w_{i,j,k}$ the weights of the
$(i,j,k)$ voxel.  The resulting images are smoothed with a Gaussian
kernel with a standard deviation of $2$ voxels, to mimic the
correlation structure observed in real fMRI data.
To simulate the spatial variability between images (inter-subject variability,
movement artifacts in intra-subject variability), we define a new support of
the ROIs, called $\tilde{\mathcal{R}}$ such as, for each image $l^{th}$, $50\%$
 (randomly chosen) of the weights $\bold{w}$ are set to zero. Thus, we have
$\tilde{\mathcal{R}}
\subset \mathcal{R}$.
We simulate the target $\bold{y}$ for the $l^{th}$ image as:
\begin{equation}
y_{l} = \sum_{(i,j,k) \in \tilde{\mathcal{R}}} w_{i,j,k} X_{i,j,k,l} +
\epsilon_l
\label{eq:sim1}
\end{equation}
with the signal in the $(i,j,k)$ voxel of the $l^{th}$ image simulated as:
\begin{equation}
X_{i,j,k,l} \sim \mathcal{N}(0,1)
\label{eq:sim}
\end{equation}
and  $\epsilon_l \sim \mathcal{N}(0,\gamma)$ is a Gaussian noise with
standard deviation $\gamma > 0.$
We choose $\gamma$ in order to have a signal-to-noise ratio of $5$
dB.
We compare TV regression cross-validated with
different values of $\lambda$ in the range $\{0.01,0.05,0.1,\}$,
with the two reference
algorithms, \emph{elastic net} and \emph{SVR}. All three methods are optimized
by 4-folds cross-validation in the range described below.

\subsection{Details on real data}

We apply the different methods on a real fMRI dataset related to an
experiment studying the representation
of objects, on ten subjects, as detailed in \cite{eger2007}.
During this experiment, ten healthy volunteers viewed objects of two categories
(each one of the two categories used in equal halves of subjects)
with $4$ different exemplars each shown in $3$ different sizes (yielding $12$
different experimental conditions), with 4 repetitions of
each stimulus in each of the 6 sessions. We pooled data from the $4$
repetitions, resulting in a total of $n=72$ images by subject (one image of
each stimulus by session).
Functional images were acquired on a 3-T MR system with eight-channel
head coil (Siemens Trio, Erlangen, Germany) as T2*-weighted
echo-planar image (EPI) volumes. Twenty transverse slices were
obtained with a repetition time of 2~s (echo time, 30~ms; flip angle,
$70^{\circ}$; $2\times2\times2$-mm voxels; $0.5$-mm gap).
Realignment, normalization to MNI space, and General Linear Model (GLM)
fit were performed with the SPM5 software
(http://www.fil.ion.ucl.ac.uk/spm/software/spm5).
The normalization is the conventional one of SPM (implying affine and
non-linear transformations) and not the one using unified segmentation. The
normalization parameters are estimated on the basis of a whole-head EPI acquired
in addition, and are then applied  to the partial EPI volumes. The data are not
smoothed.
In the GLM, the effect of each of the $12$ stimuli convolved with a standard
hemodynamic response function was modeled separately, while
accounting for serial autocorrelation with an AR(1) model and
removing low-frequency drift terms by a high-pass filter with a
cut-off of 128 s.
The GLM is fitted separately in each session for each subject, and we used
in the following analyzes the resulting session-wise parameter estimate images
the $\beta$-maps are used as rows of $\bold{X}$).
All the analyzes are performed without any prior selection of regions of
interest, and use the whole acquired volume.

\begin{figure}[htb]
\center   \includegraphics[width=0.8\linewidth]{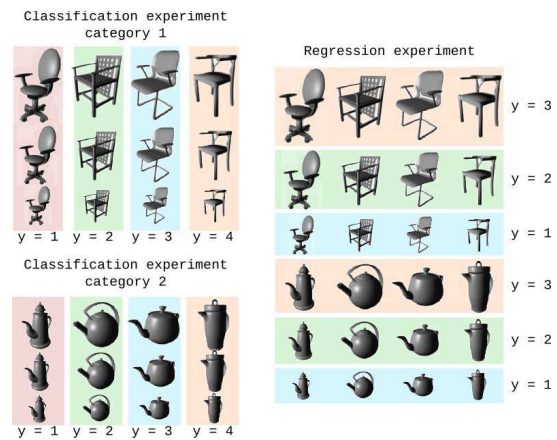}
\caption{Experiment paradigm for the classification of
object in each of the category (left) and regression (right) experiments. Each
color represents the stimuli which are pooled together in one of the three
experiments (classification category 1, classification category 2 and
regression).}
\label{Fig:design}
\end{figure}
\textbf{Regression experiments:} First, we perform an intra-subject regression
analysis.
The four different shapes of objects (for the two categories)
were pooled across for each one of the three sizes, and we are
interested in finding discriminative information between sizes. This
reduces to a regression problem, in which our goal is to predict a
simple scalar factor (size of an object) (see Fig.~\ref{Fig:design}).
Each subject is evaluated independently, in a 12-fold
cross-validation.  The dimensions of the real data set for one subject
are $p \sim 7\times 10^4$ and $n=72$ (divided in $3$ different sizes,
$24$ images per size). We evaluate the performance of the method by a
leave-one-condition-out cross-validation (\emph{i.e.},
leave-6-images-out), and doing so the GLM is performed separately for the
training and test sets.
The parameters of the reference methods are optimized with a nested
leave-one-condition-out cross-validation within the training set, in
the ranges given before.
Additionally, we perform an inter-subject regression analysis on the
sizes.  The inter-subject analysis relies on subject-specific
fixed-effects activations, \emph{i.e.} for each condition, the $6$ activation
maps corresponding to the $6$
sessions are averaged together.  This yields a total of $12$ images
per subject, one for each experimental condition. The dimensions
of the real data set are $p \sim 7\times 10^4$ and $n=120$ (divided in
$3$ different sizes).  We evaluate the performance of the method by
cross-validation (leave-one-subject-out).
The parameters of the reference methods are optimized with a
nested leave-one-subject-out cross-validation within the training set, in the
ranges given before.

\textbf{Classification experiments:} We evaluate the performance on a second
type of discrimination which is object classification (see
Fig.~\ref{Fig:design}). In that case, we averaged the images for the three
sizes and we are interested in discriminating between individual object
exemplars/shapes.  For each of the two categories, this can be handled
as a classification problem, where we aim at predicting the shape of
an object corresponding to a new fMRI scan.  In order to investigate the
performance of TV classification, which is an original contribution, we
perform an inter-subject analysis in the same way as described for the
regression study, except that now, we perform two analyzes corresponding to the
two categories used, each one including $5$ subjects.

\begin{figure}[h!tb]
\center\includegraphics[width=1.\linewidth]{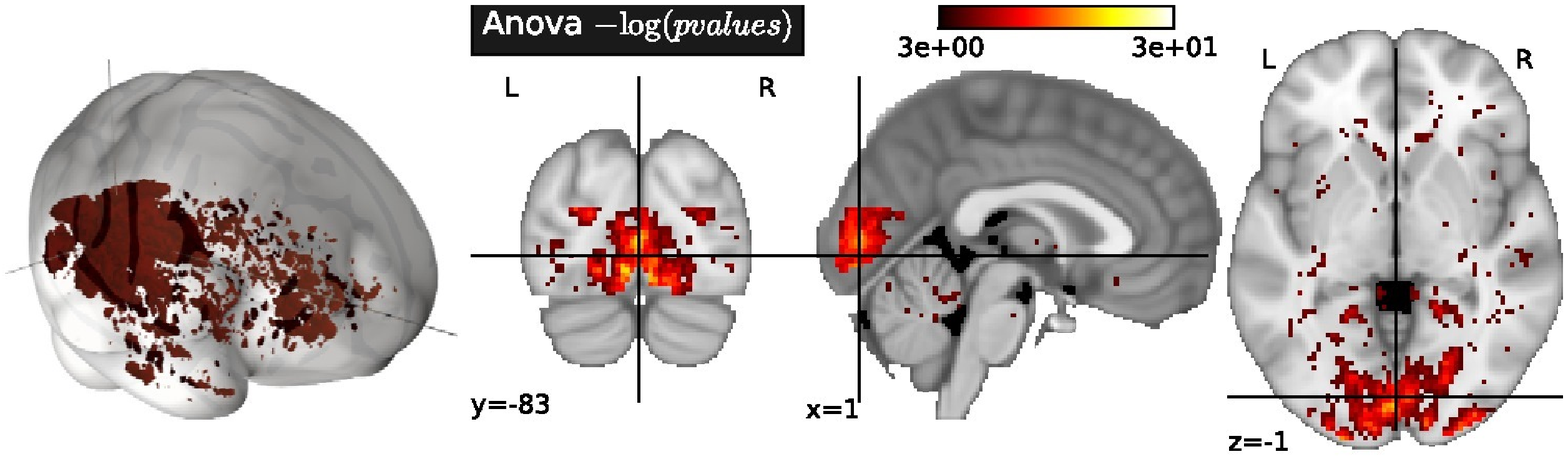}
\center\includegraphics[width=1.\linewidth]{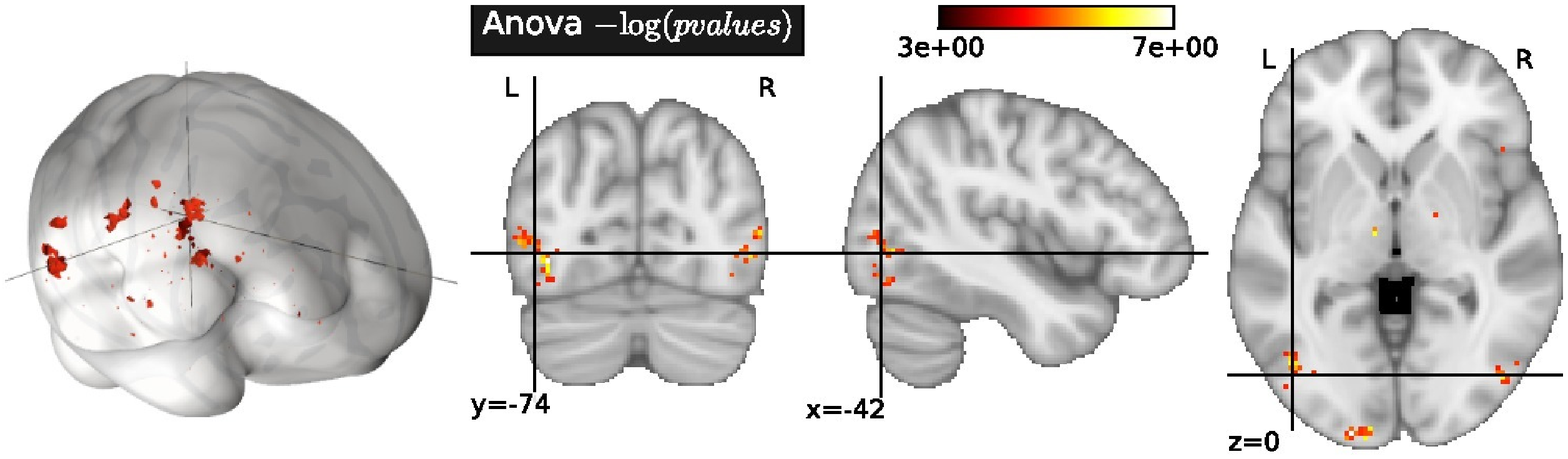}
\center\includegraphics[width=1.\linewidth]{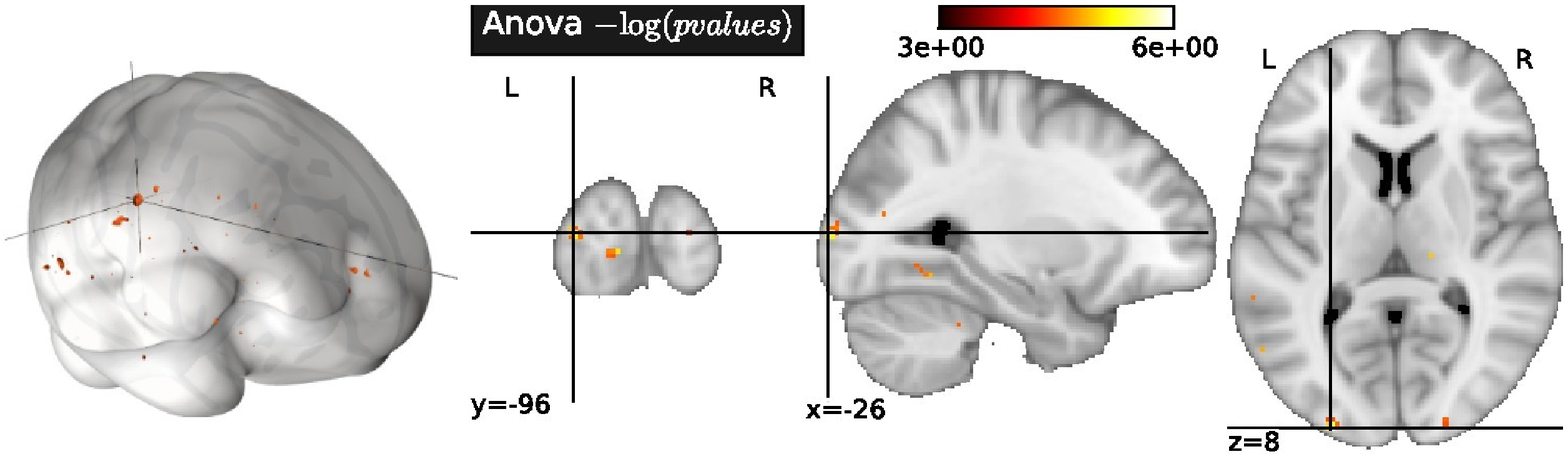}
\vspace{-4ex}
  \caption{Inter-subject analysis - Maps of \emph{Anova}
($-\log(\textrm{p-values})$) for the sizes prediction experiment (top) and the
objects identifications for category 1 (middle) and category 2 (bottom). We
threshold  the p-values higher than $10^{-3}$ (i.e. $-\log(\textrm{p-values}) >
3$).}
\label{Fig:real_anova}
\end{figure}

\textbf{Statistical Parametric Maps:}
For comparison purposes, the corresponding maps of \emph{Anova}
(\emph{F-score}), or \emph{SPMs}, for
the inter-subject analysis are given Fig.~\ref{Fig:real_anova}, for the
representation of sizes (top) and representation of objects for
the two categories (middle  and bottom).
As expected, the sizes are mostly processed in primary visual
cortex, while for objects, discrimination is additionally observed in lateral
occipital regions \cite{eger2007}.

\section{Results}
\label{sec:results}

\subsection{Results on simulated data}

\begin{figure*}[h!tb]

\begin{center}
\begin{minipage}{0.19\linewidth}
\center \includegraphics[width = 0.7\linewidth]{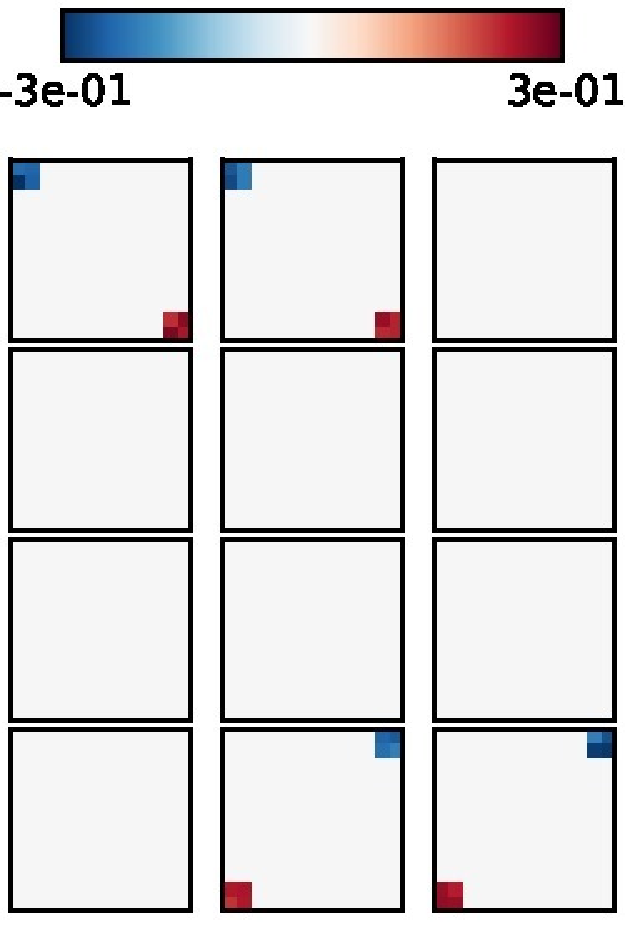}
\center \vspace{-3ex} \emph{(a) True weights}
\end{minipage}
\begin{minipage}{0.19\linewidth}
\center \includegraphics[width = 0.7\linewidth]{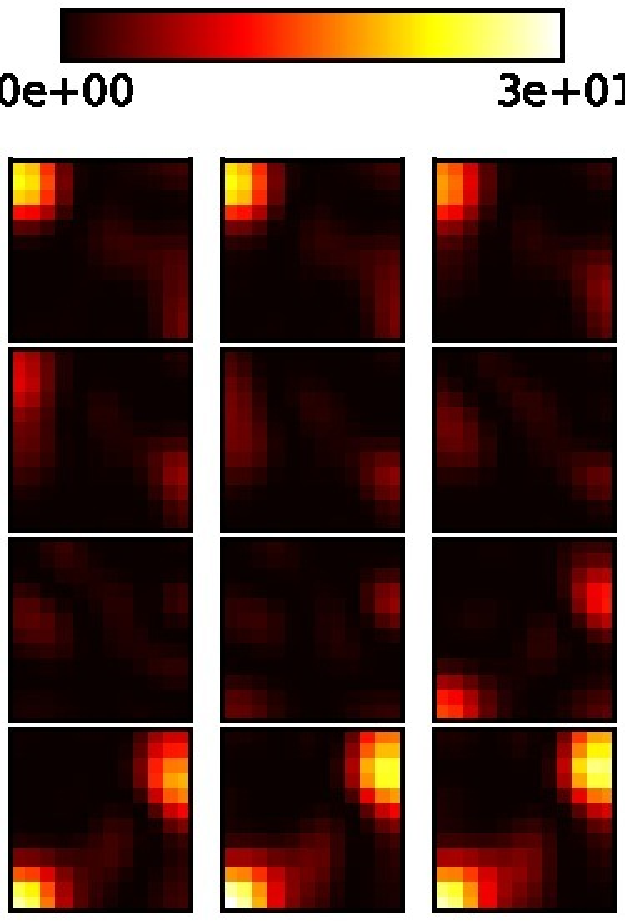}
\center \vspace{-3ex} \emph{(b) F-scores}
\end{minipage}
\begin{minipage}{0.19\linewidth}
\center \includegraphics[width = 0.7\linewidth]{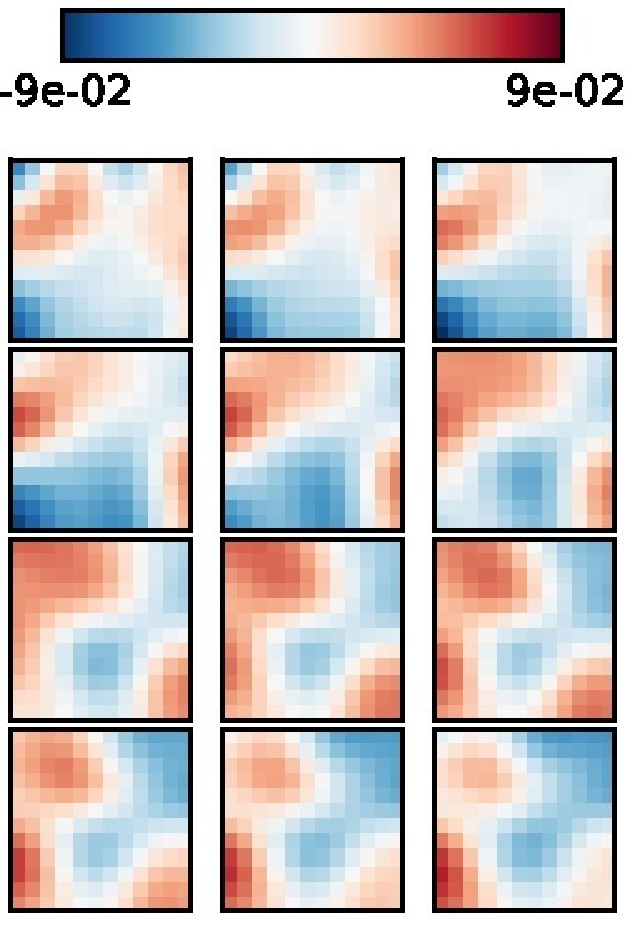}
\center \vspace{-3ex} \emph{(c) C.v. SVR}
\end{minipage}
\begin{minipage}{0.19\linewidth}
\center \includegraphics[width = 0.7\linewidth]{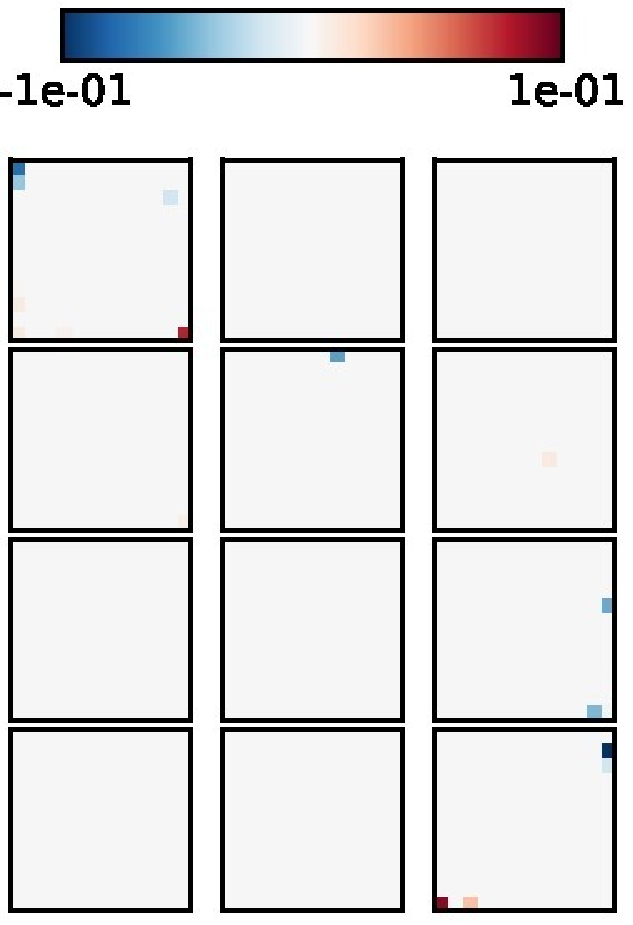}
\center \vspace{-3ex} \emph{(d) C.v. Elastic net}
\end{minipage}
\begin{minipage}{0.19\linewidth}
\center \includegraphics[width = 0.7\linewidth]{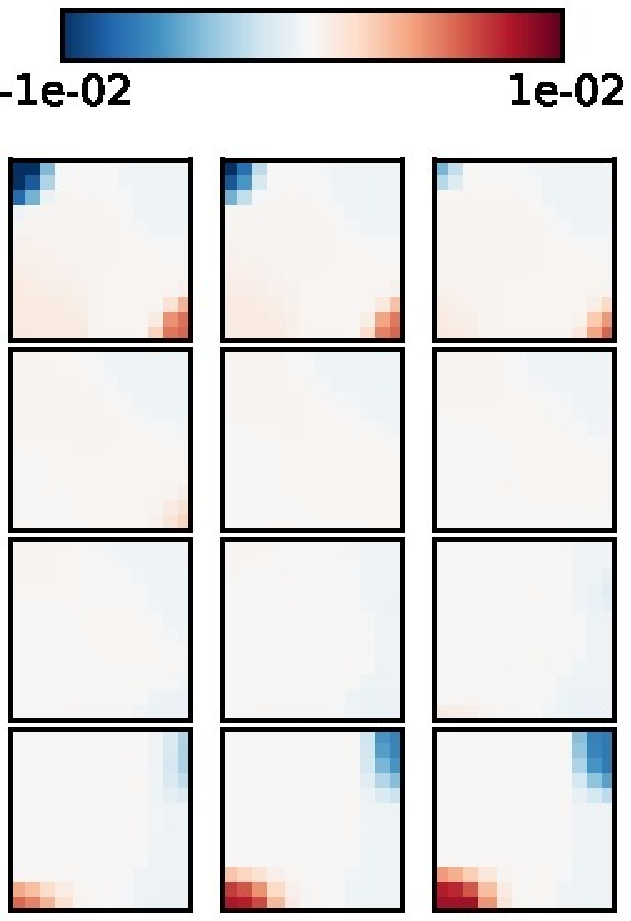}
\center \vspace{-3ex} \emph{(e) C.v. TV}
\end{minipage}
\end{center}

\vspace{-3ex}

\begin{center}
\begin{minipage}{0.19\linewidth}
\center \includegraphics[width = 0.6\linewidth]{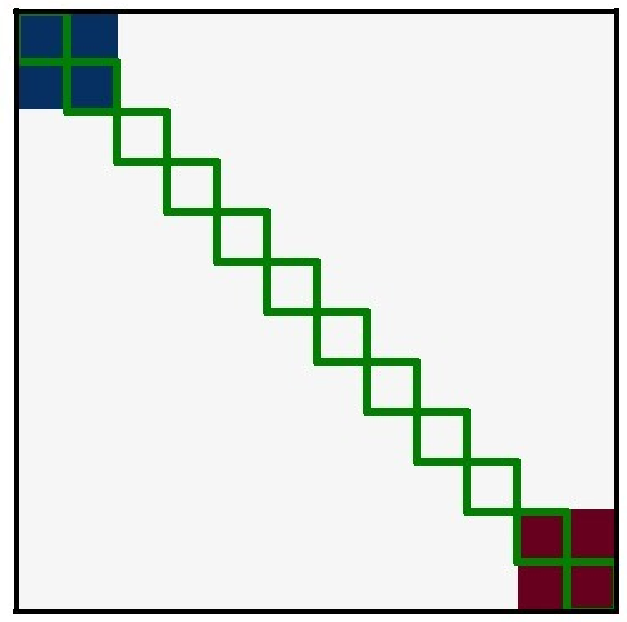}
\center \vspace{-3ex} \emph{Slice}
\end{minipage}
\hfill
\begin{minipage}{0.19\linewidth}
\center \includegraphics[width = 1.\linewidth]{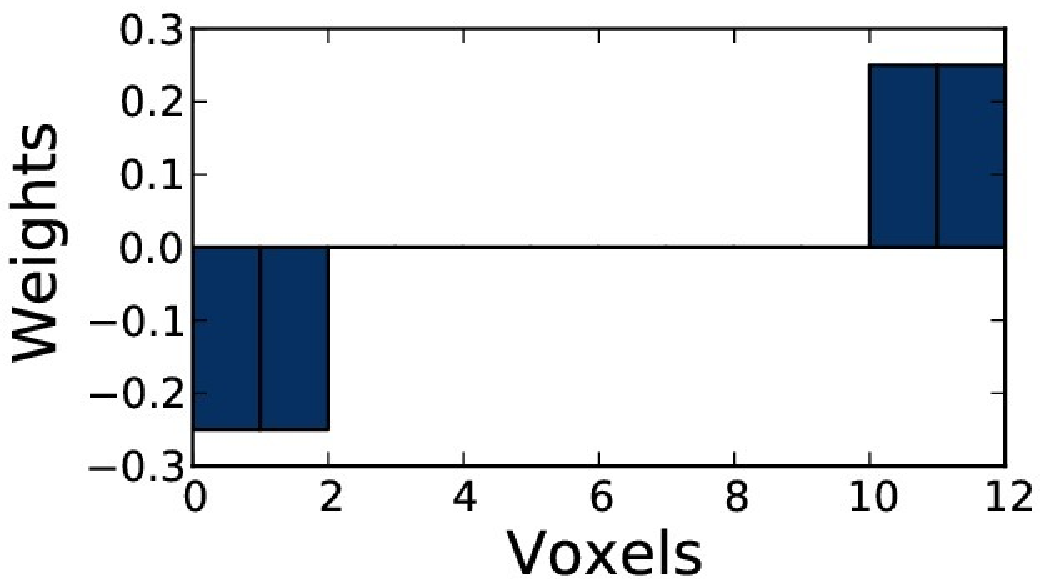}
\center \vspace{-2ex} \emph{(a) True weights}
\end{minipage}
\hfill
\begin{minipage}{0.19\linewidth}
\center \includegraphics[width = 1.\linewidth]{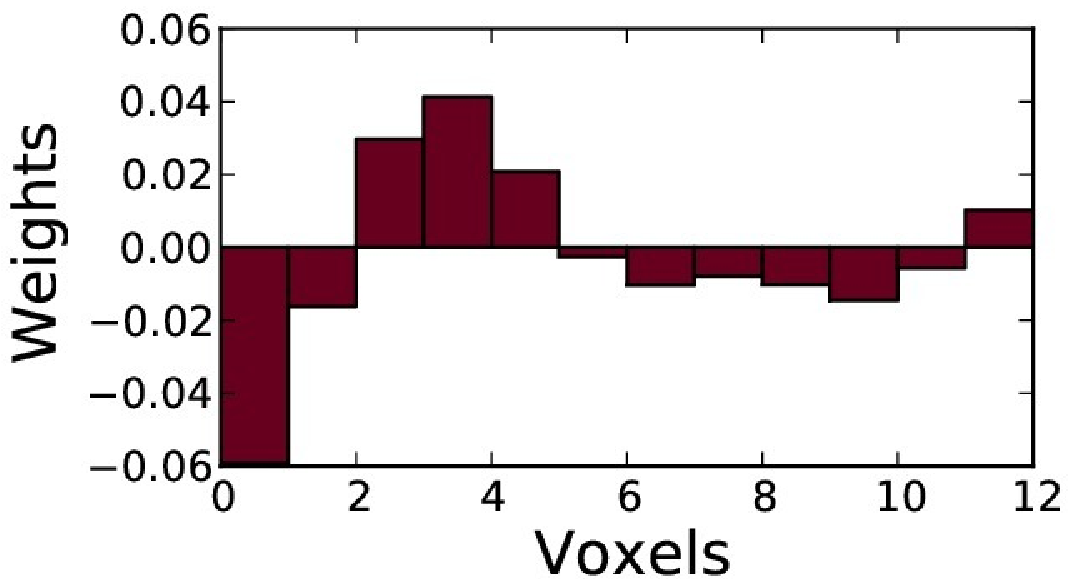}
\center \vspace{-2ex} \emph{(c) C.v. SVR}
\end{minipage}
\hfill
\begin{minipage}{0.19\linewidth}
\center \includegraphics[width = 1.\linewidth]{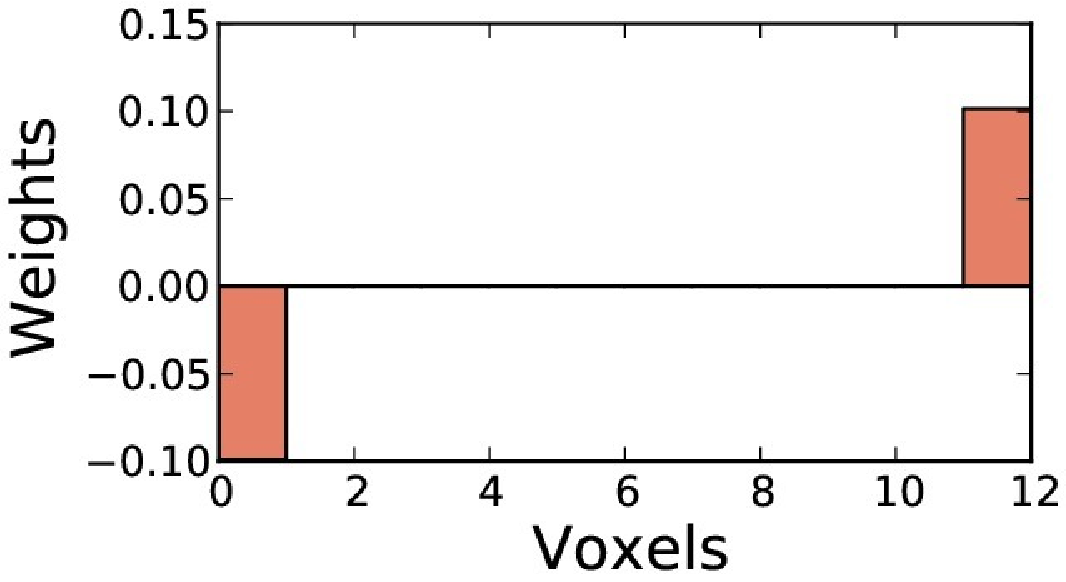}
\center \vspace{-2ex} \emph{(d) C.v. Elastic net}
\end{minipage}
\hfill
\begin{minipage}{0.19\linewidth}
\center \includegraphics[width = 1.\linewidth]{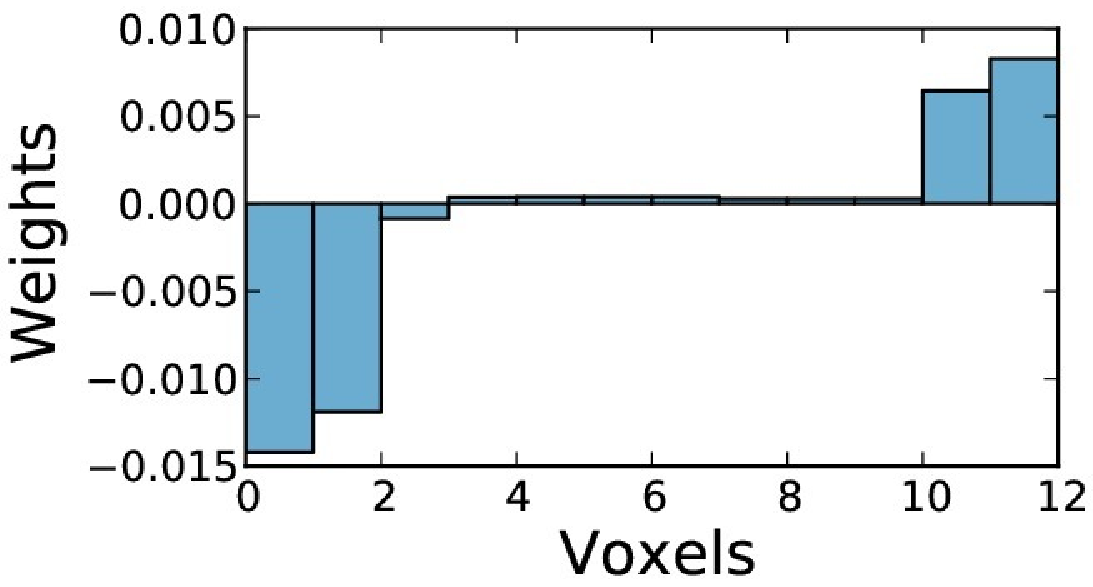}
\center \vspace{-2ex} \emph{(e) C.v. TV}
\end{minipage}
\end{center}

\vspace{-3ex}

\begin{center}
\caption{
Two-dimensional slices of the three-dimensional volume of simulated
data (top), and weights found on the diagonal (green squares) of the
first two-dimensional slice (bottom). Comparisons of the weights found
by different methods, with the true target (a), and the \emph{F-score}
found by \emph{Anova} (b).  The TV method (e) retrieves the
true weights.  The reference methods ((c), (d)) yield less accurate
maps. Indeed, the support of the weights found by \emph{elastic net}
is too sparse and does not yield convex regions. \emph{SVR} yields
smooth maps that do not look like the ground truth.}
\label{fig:ResSimuParam}
\end{center}

\vspace{-4ex}

\end{figure*}

We compare the different methods on the simulated data, see
Fig.~\ref{fig:ResSimuParam}. The true weights (a) and resulting
\emph{Anova} F-scores (b) are shown.  Only TV regression (e) is
able to extract the simulated discriminative regions. \emph{Elastic net} (d)
only
retrieves part of the support of the weights, and yields an overly
sparse solution.
This sparsity pattern obtained with \emph{elastic net} is the one that
yields the highest prediction accuracy: one could seek a less sparse solution,
but this would decrease the prediction accuracy.
We note that the weights in the primal space estimated by \emph{SVR} (c) are
everywhere non-zero and do not retrieve the support of the weights.

\subsection{Sensitivity study on real data}

\begin{figure}[h!tb]
\begin{center}
\includegraphics[width=1.\linewidth]{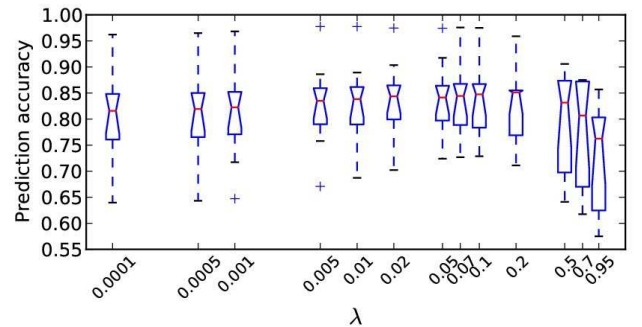}
\end{center}
\begin{center}
\vspace{-3ex}
\caption{
Explained variance $\zeta$ for different values of $\lambda$, in the
inter-subjects regression analysis. The accuracy is very stable regarding to
$\lambda$ in the range $[5.10^{-4},5.10^{-1}]$.}
\label{Fig:sensitivity}
\end{center}
\end{figure}

Before any further analysis on real data, we have performed a sensitivity
analysis of our model with regards to the parameter $\lambda$.
In the inter-subject analysis for the size regression, we compute the
cross-validated prediction accuracy for twelve different values of $\lambda$
between $10^{-4}$ and $0.95$.
The aim of the sensitivity study is to assess the stability of the
prediction with respect to the regularization parameter. The results, detailed
in Fig.~\ref{Fig:sensitivity}, are extremely stable with respect to
$\lambda$ in the range $[5.10^{-4},5.10^{-1}]$. For this reason, we can
fix $\lambda$ = 0.05 in the following analyzes.
The value of $\lambda$ is the same for all the experiments, in both
classification and regression settings. 
The correct way of choosing the regularization parameter is to
embed the TV regularization within an internal cross-validation on the training
set. However, such approach can be computationally costly.

\subsection{Results for regression analysis}

In a first set of analyzes, we assess the performance of TV regression in
both intra-subject and inter-subject cases, where the
aim is to predict the size of an object seen by the subject during the
experiment.

\subsubsection*{Intra-subject analysis} The results obtained by the three
methods are given in Tab.~\ref{Fig:res_sizes_intra_reg}. TV
regression outperforms the two
alternative methods, yielding an average explained variance of $0.92$ across the
subjects. The difference with \emph{SVR} is significant, but not with
\emph{elastic net}.
Moreover, the results of the regularized methods (\emph{TV, elastic
net}) are more stable (standard deviation three times smaller) across
subjects, than the results of the \emph{SVR}.

\begin{table}[h!tb]
\begin{center}
\footnotesize{
\begin{tabular}{|l|c|c|c|c|c|}
\hline
 Methods & mean $\zeta$ & std $\zeta$ & max $\zeta$ & min $\zeta$ & p-value to
TV  \\
\hline
\hline
SVR & $0.82$ & $0.07$ & $0.9$ & $0.67$ & $0.0051$\\
\hline
Elastic net & $0.9$ & $0.02$ & $0.93$ & $0.85$ & $0.0745$ \\
\hline
TV $\alpha=0.05$ & $0.92$ & $0.02$ & $0.95$ & $0.88$ & - \\
\hline
\end{tabular}
}
\end{center}
\caption{\emph{Regression - Sizes prediction experiment - Intra-subject
analysis}.
Explained variance $\zeta$ for the three different methods.
TV regression yields the best prediction accuracy, while
being more stable than the two reference methods (standard deviation of
$\zeta$  three times smaller than \textit{SVR}). The p-values are computed
on the explained variance of the different subjects.}
\label{Fig:res_sizes_intra_reg}

\end{table}

\subsubsection*{Inter-subject analysis}
The results obtained with the three methods are given in
Tab.~\ref{Fig:res_sizes_inter_reg}. As in the intra-subject analysis,
\emph{TV regression} outperforms the two alternative methods, yielding
an average explained variance of $84\%$, and also more stable
predictions.  Such stability can be illustrated on the subject $3$,
where both reference methods yield poor results, while TV
regression yields an explained variance $0.2$ higher. Moreover, we have
tested that feature selection minimizes
overfitting. Indeed, without such feature selection, we obtain a smaller
explained variance of $76\%$ for \emph{SVR} and $64\%$ for \emph{elastic net}.

\begin{table}[h!tb]
\begin{center}
\footnotesize{
\begin{tabular}{|l|c|c|c|c|c|}
\hline
 Methods & mean $\zeta$ & std $\zeta$ & max $\zeta$ & min $\zeta$ & p-value to
TV  \\
\hline
\hline
SVR & $0.77$ & $0.11$ & $0.97$ & $0.58$ & $0.0284$\\
\hline
Elastic net & $0.78$ & $0.1$ & $0.97$ & $0.65$ & $0.0469$\\
\hline
TV $\lambda=0.05$ & $0.84$ & $0.07$ & $0.97$ & $0.72$ & - \\
\hline
\end{tabular}
}
\end{center}
\caption{\emph{Regression - Sizes prediction experiment - Inter-subject
analysis}.
Explained variance $\zeta$ for the three different methods.
TV regression still yields the best prediction accuracy, with an
explained variance $0.06$ higher than the best reference method (\emph{elastic
net}). The p-values are computed on the explained variance of the different
subjects.}
\label{Fig:res_sizes_inter_reg}

\end{table}

\begin{figure}[h!tb]
       \center   \includegraphics[width=1.\linewidth]{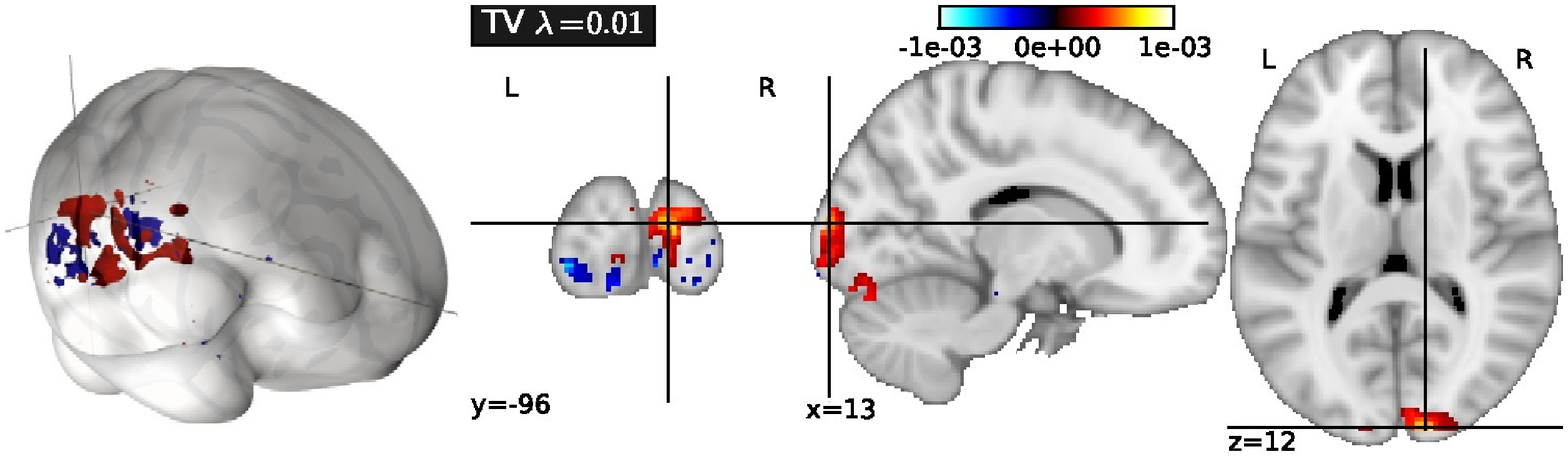}
  \vspace{-10ex}\center   $$\zeta = 0.83$$
       \center   \includegraphics[width=1.\linewidth]{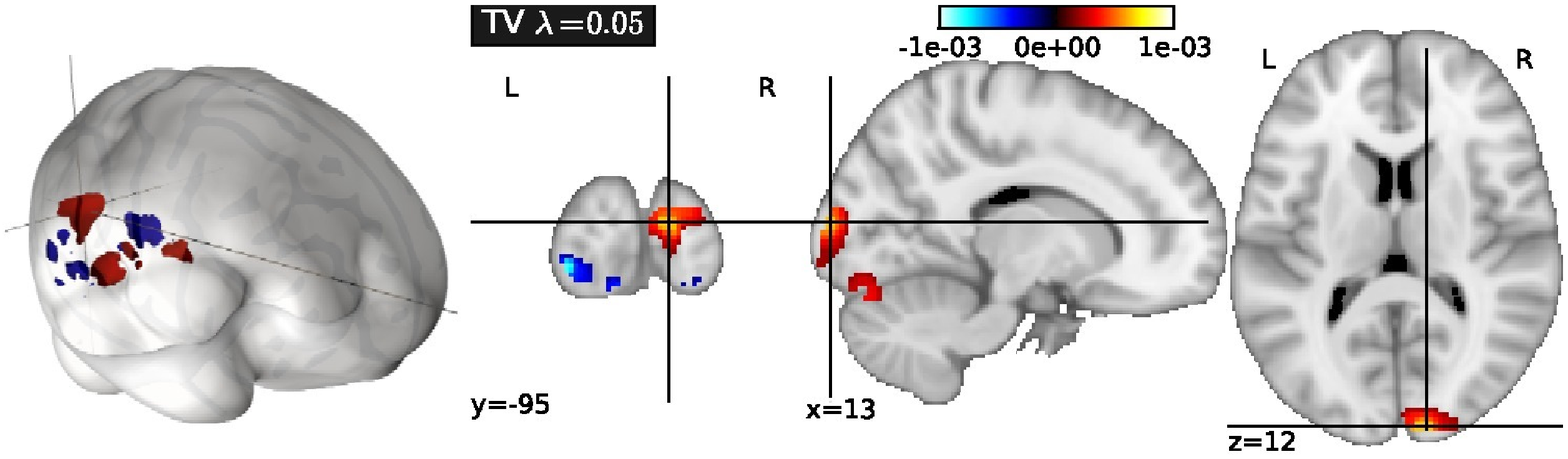}
   \vspace{-10ex}\center   $$\zeta = 0.84$$
       \center   \includegraphics[width=1.\linewidth]{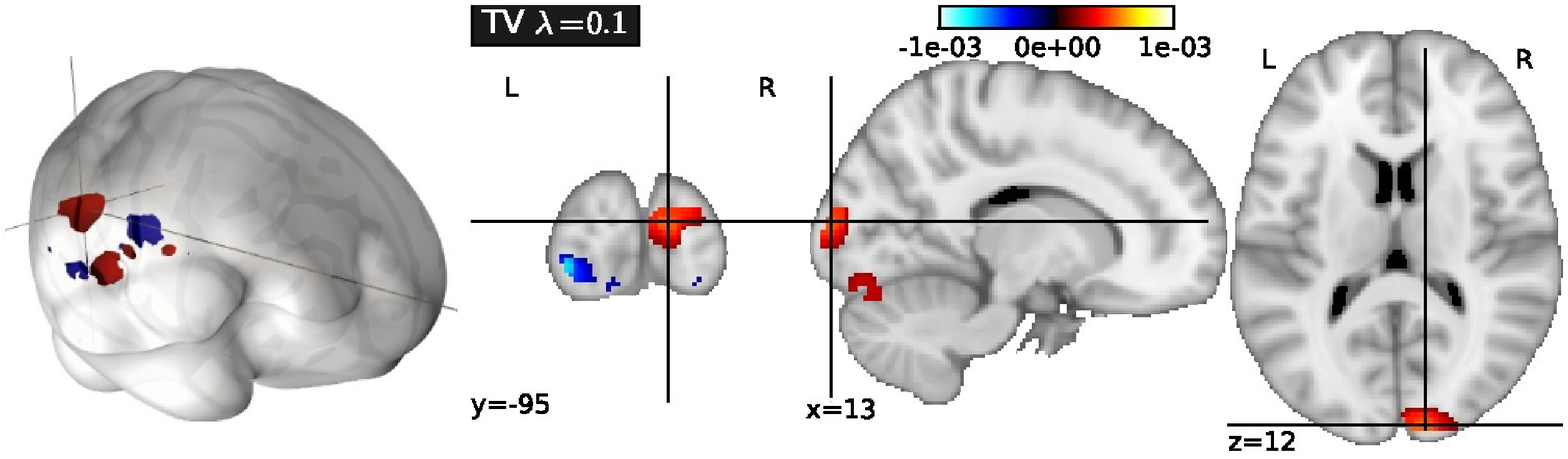}
   \vspace{-10ex}\center   $$\zeta = 0.84$$
\vspace{-5ex}
\caption{\emph{Regression - Sizes prediction experiment - Inter-subject
analysis}.
Maps of weights found by TV regression for various
    values of the regularization parameter $\lambda$. When $\lambda$
    decreases, the TV regression algorithm creates different clusters
    of weights with constant values. These clusters are easily
    interpretable, compared to voxel-based map (see
    below). The TV regression algorithm is very stable for
    different values of $\lambda$, has shown by the explained variance
    $\zeta$.}
  \label{Fig:real_w_sizes_inter}
  \vspace{2ex}
\center   \includegraphics[width=1.\linewidth]{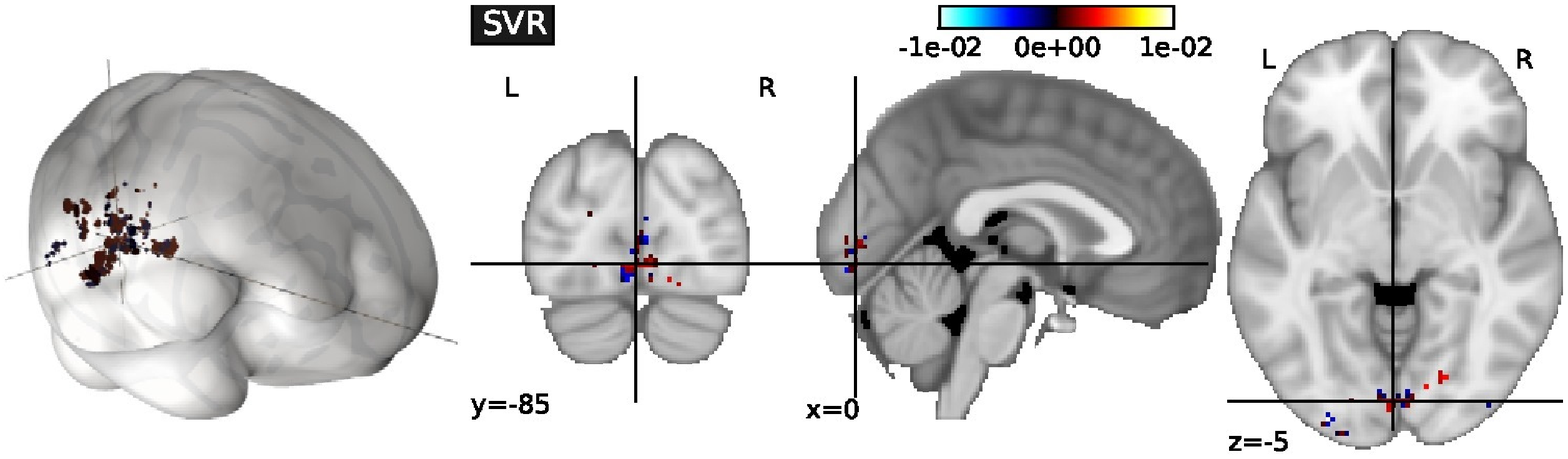}
\vspace{-10ex}\center  $$\zeta = 0.77$$
       \center   \includegraphics[width=1.\linewidth]{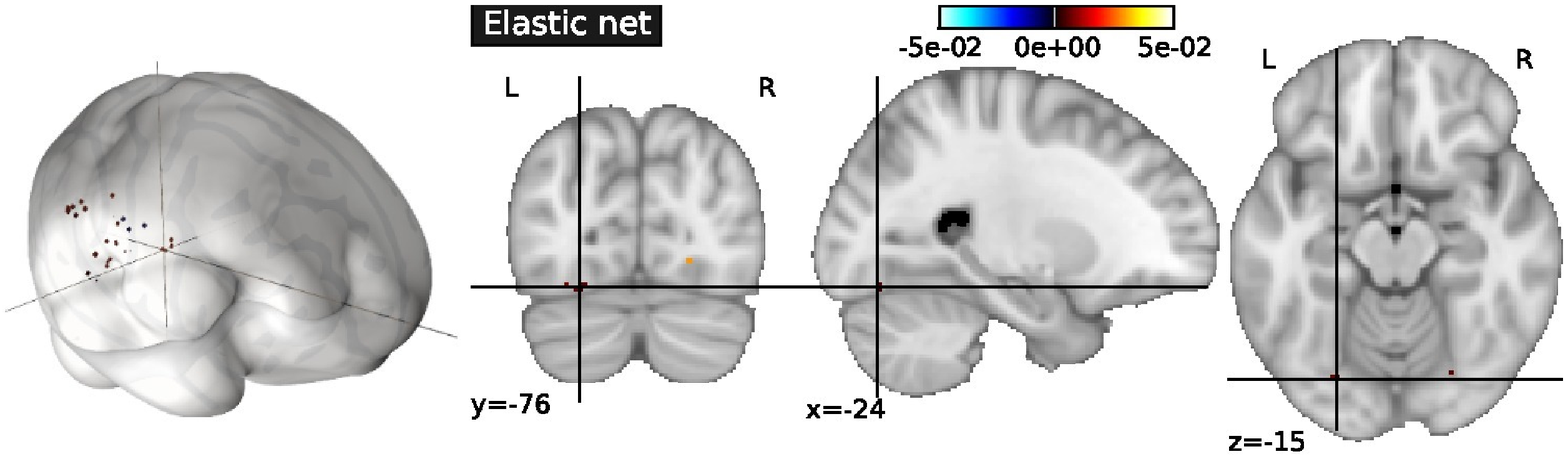}
\vspace{-10ex}\center  $$\zeta = 0.78$$
\vspace{-5ex}
  \caption{\emph{Regression - Sizes prediction experiment - Inter-subject
analysis}.
    Maps of weights found by the \textit{SVR} (top) and
    \textit{elastic net} (bottom) methods. The optimal number of
    voxels selected by \emph{Anova} is $500$, but \textit{elastic net} further
    reduces this set to 21 voxels.  These voxel-based methods do not
    yield interpretable features (especially when compared to TV 
    regression), which is due to the fact that they do not consider the
    spatial structure of the image.}
\label{Fig:real_w_ref_sizes_inter}
\end{figure}

The weighted maps found by the different methods are given in
Fig.~\ref{Fig:real_w_sizes_inter}.  One can notice that, as $\lambda$
increases, the spatial support of these maps tends to be aggregated in
few clusters within the occipital cortex, and that the maps have a
nearly constant value on these clusters.  By contrast, both reference
methods yield uninterpretable (\emph{i.e.} more complex) maps, with a
few informative voxels scattered in the whole occipital cortex.
The average positions and the sizes of the three main clusters found by the
TV algorithm, using all the subjects, are given
Tab.~\ref{Fig:clusters_inter_reg}.
TV regression is able to adapt the
regularization to tiny regions, yielding ROIs from $25$ to
$193$ voxels.  The clusters are found within the occipital cortex.
The majority of informative voxels are located in the posterior part of
the occipital cortex ($y\leq-90$ mm),
most likely corresponding to
primary visual cortex, with one additional slightly more anterior
cluster in posterior lateral occipital cortex. This is consistent with
the previous findings \cite{eger2007} where a gradient of sensitivity
to size was observed across object selective lateral occipital ROIs,
and the most accurate discrimination of sizes in primary visual
cortex.

\begin{table}[h!tb]
\begin{center}
\footnotesize{
\begin{tabular}{|c|c|c|c|}
\hline
 x (mm)& y (mm)& z (mm)& Sizes (voxels)\\
\hline
\hline
24 &-92 &-16 &25 \\
-26 &-96 &-10 &103 \\
16 &-96 &12 &193 \\
\hline
\end{tabular}
}
\end{center}
\caption{Inter-subject regression analysis: positions and sizes of the three
main clusters for the TV regression method.}
\label{Fig:clusters_inter_reg}

\end{table}

\subsection{Results of classification experiments}

In a second analysis, we assess the performance of TV classification in
an inter-subject classification analyzes, in which the aim is
to predict which of $4$ object exemplars is seen by the different subjects.

\begin{figure}[h!tb]
\center \includegraphics[width=1.\linewidth]{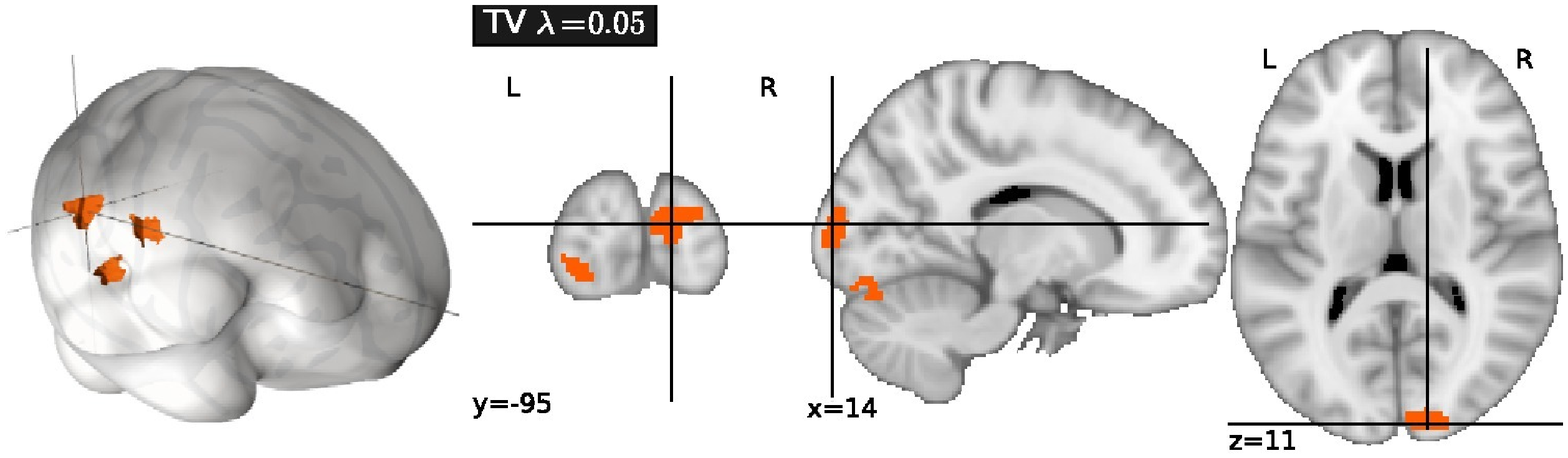}
\vspace{-4ex}
\center \includegraphics[width=1.\linewidth]{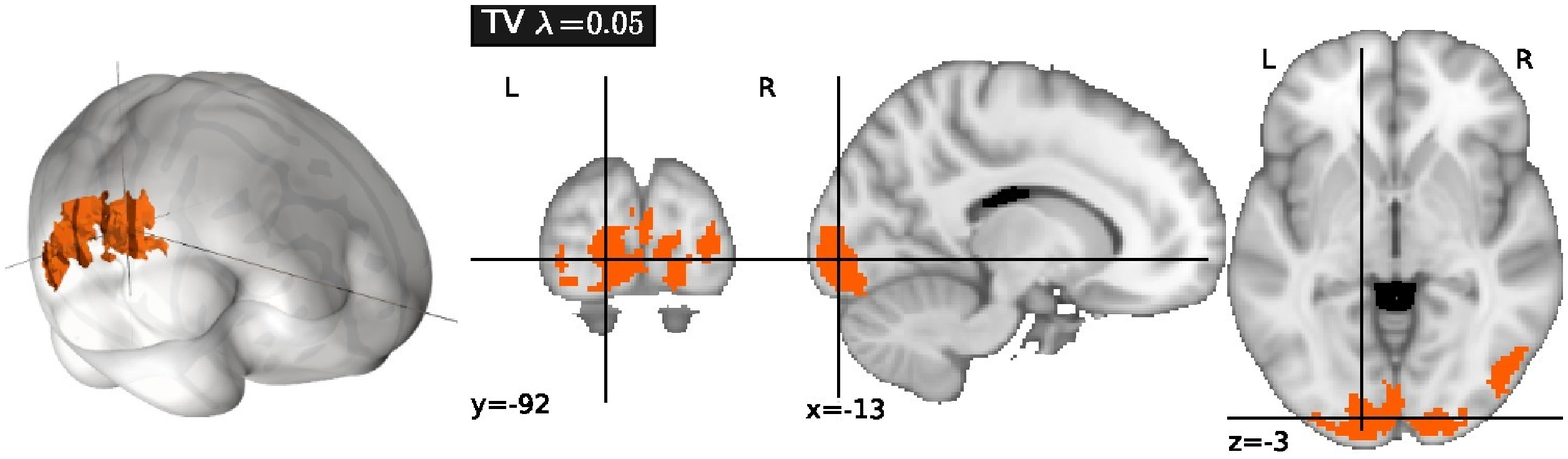}
\vspace{-4ex}
\caption{Inter-subject analysis. Top - voxels selected within one of the three
main clusters by TV regression, for the \emph{Sizes prediction
experiment}.
Bottom - voxels selected at least one time within one of the three main clusters
for each of the one-vs-one TV classification, for the \emph{Objects
prediction experiment}.
Some clusters found in the \emph{Objects prediction experiment} ($y = -40$ mm,
$y = -74$ mm)  are more anterior than the ones found for the \emph{Sizes
prediction
experiment} ($y = -92$ mm, $y = -96$ mm).
This is coherent with the hypothesis that
the processing of shapes is done at a higher level in the processing of visual
information, and thus the implied regions are found further in the ventral
pathway.
}
\label{Fig:clusters_inter}
\end{figure}

The results (average across the two categories) found by the three methods are
given in Tab.~\ref{Fig:res_objects_inter_classif}.
As in the inter-subject regression analysis, the TV-based method
outperforms the \emph{SMLR} method. Moreover, it yields an average
classification score
similar to the \emph{SVC} while being more stable.
Seeking clusters of activation thus seems a reasonable way to cope with
inter-subject variability.
The average number of selections of each
voxels within one of the three larger clusters
for each one-versus-one map are given Fig.~\ref{Fig:clusters_inter}. The
informative clusters are more anterior and more ventral than the ones found
within the sizes prediction paradigm. We thus confirm the results found by
classical brain mapping approach, such as  \emph{Anova} (see
Fig.~\ref{Fig:real_anova}), while providing a classification score based on
cross-validation on independent data which allows to
check the actual implication of these regions in the cognitive process.

\begin{table}[h!tb]
\begin{center}
\footnotesize{
\begin{tabular}{|l|c|c|c|c|c|}
\hline
 Methods & mean $\kappa$ & std $\kappa$ & max $\kappa$ & min $\kappa$ & p-value
to SVC\\
\hline
\hline
SVC & $48.33$ & $15.72$ & $75.0$ & $25.0$ & - \\
\hline
SMLR & $42.5$ & $9.46$ & $58.33$ & $33.33$ & $0.0$ **\\
\hline
TV $\lambda=0.05$ & $46.67$ & $11.3$ & $66.67$ & $25.0$ & $1.0$ \\
\hline
\end{tabular}
}
\end{center}
\vspace{-2ex}
\caption{\emph{Classification - Objects prediction experiment
}. Averaged classification score $\kappa$ for the three different
methods, across the two categories. TV classification yields similar
prediction accuracy than the reference method \emph{SVC}. The p-values are
computed on the classification score of the different
subjects.}
\label{Fig:res_objects_inter_classif}
\end{table}

\section{Discussion}

In this article, we present the first use of TV regularization
for brain decoding. This method outperforms the reference methods on
prediction accuracy, and yields sparse brain maps with clear informative foci.

Moreover, with regard to the classification paradigm, we integrate the
TV in a logistic regression framework. This approach, which to
our knowledge, has not been used before, yields high prediction
accuracy, and seems to be a promising method for more machine learning
problems beyond the scope of neuroimaging.

One major advantage of the proposed method is that, in the case of a
multi-subject studies, considering extended regions is expected to
compensate for spatial misalignment, hence it can better generalize than
voxel-based methods. As proven on both inter-subject analyzes, the
proposed TV approach yields significantly higher prediction
accuracy than reference voxel-based methods.
In addition, the proposed approach yields weight maps very similar to the
maps obtained by a classical brain mapping approach (such as
\emph{Anova}).
We note that the solution found by our method has a sparse
block structure and is sufficient for good prediction accuracy, which explains
the fact
that the regions observed may be more compactly localized than the ones from
\emph{Anova}.
Thus, the TV approach has the assets of a predictive
framework, while leading to accurate brain maps. It is important to notice that,
even if TV does not promote a strict sparsity of the weights, most voxels
are associated with very small weights, and only a few clusters get high
weightings.
Moreover, TV regression allows to consider the whole brain in
the analysis, without requiring any prior feature selection. As many
accurate dimension reduction approaches such as \emph{Recursive
Feature Elimination} \cite{guyon2002} can be extremely costly in
computational time, avoiding this step is a major asset.
An important feature of our implementation is thus that it reduces
computation time to a reasonable amount, so that it is not
significantly more costly than SVR or elastic net in practical
settings (\textit{i.e.}, including the cross-validation loops).  In
the inter-subject regression analysis, the average computational time
is $185$ seconds for TV regression, $131$ seconds for
\emph{Anova + SVR} and $121$ seconds for \emph{Anova + elastic net},
on a \emph{Intel(R) Xeon(R) CPU} at $2.83$GHz.
Regularization of the voxel weights significantly increases the
generalization ability in regression problems, by performing feature
selection and training of the prediction function jointly.
However, to date, regularization has most often been performed without
using the spatial structure of the images.
By applying a penalization on the gradient of the weight and thus
taking into account the spatial structure of the image-based
information, our approach performs an adaptive and efficient
regularization, while creating sparse weight maps with regions of
quasi constant weights.
TV regularization method fulfills thus the two requirements that
make it suitable for neuroimaging brain mapping: a good prediction
accuracy (better than the reference methods for regression experiments, and
equal for classification),
and a set of interpretable features, made of clusters of
similarly-tuned voxels. In that sense, it can be seen as the first
method for performing a large scale multivariate brain mapping 
(the searchlight \cite{kriegeskorte2006} only consider the multivariate
information in a small neighborhood).

From a neuroscientific point of view, the regions extracted from the whole
analysis volume in the size discrimination task are concentrated
in the early visual cortex.
This is consistent with the fact that early visual cortex yields
highly reliable signals that are discriminative of feature/shape
differences between object exemplars, which holds as long as no
high-level generalization across images is required (see \emph{e.g.}
\cite{cox2003} and \cite{eger2007}).
This is expected given the small receptive fields of neurons in these
regions that will reliably detect differences in the spatial envelop
or other low-level structure of the images.
Most importantly, the predictive spatial pattern is stable enough
across individuals to make reliable predictions in new subjects. In
fact our method compares best with regards to the state of the art in
the inter-subjects setting, because it selects predictive regions that are
not very sensitive to anatomo-functional variability.
In the object discrimination task, the clusters found by our approach
are also in the visual cortex, but including more anterior ones
(probably corresponding to posterior \emph{lateral occipital} region)
compared to size discrimination, which is consistent with the fact
that shape discrimination requires intermediate/higher level visual
areas. The finding that also large parts of early visual cortex were
discriminative here is explained by the fact that generalization
across viewing conditions was not a part of the analysis and
classification can therefore be driven by lower-level features.
However, even if similar maps as the ones found by our method can be
obtained using \emph{Anova}, they do not provide a prediction score
for generalization to independent data (\emph{i.e.} a global measure of the
involvement of the regions in the cognitive process).

\section{Conclusion}
In this paper we introduce TV regularization for extracting
information from brain images, both in regression or classification settings.
Feature selection and model estimation are performed jointly and
capture the predictive information present in the data better than
alternative methods.
A particularly important property of this approach is its ability to
create spatially coherent regions with similar weights, yielding simplified and
still informative sets of features.
Experimental results show that this algorithm performs well on real
data, and is far more accurate than voxel-based reference methods for
multi-subject analysis. In particular, the segmented regions are
robust to inter-subject variability.
These observations demonstrate that TV regularization is a
powerful tool for understanding brain activity and spatial mapping of
cognitive process, and is the first method that is able to derive
statistical weight maps, as in the standard \emph{SPM} approach,
within the inverse inference framework.

\medskip
\noindent
\textbf{Acknowledgments:} The authors acknowledge support from the ANR grant
ViMAGINE ANR-08-BLAN-0250-02.

\bibliographystyle{IEEEtran}
\bibliography{IEEEabrv,bib_tv}

\appendices
\section{Gradient and Divergence}
\label{appendix_grad}

The gradient operator which has to be computed
on a mask in our case (mask of the brain).
With $I \in \mathbb{R}^{p_i \times p_j \times p_k}$ an image, it
is defined by:
\begin{eqnarray*}
(\grad\;I)_x^{i,j,k} =
\begin{cases}
    I_{i+1,j,k} - I_{i,j,k} &\mbox{if } M_{i,j,k} = M_{i+1,j,k} = 1
\\
    0 & \mbox{otherwise}
\end{cases}
\\
(\grad\;I)_y^{i,j,k}=
\begin{cases}
    I_{i,j+1,k} - I_{i,j,k} &\mbox{if } M_{i,j,k} =  M_{i,j+1,k} =1
\\
    0 & \mbox{otherwise}
\end{cases}
\\
(\grad\;I)_z^{i,j,k}=
\begin{cases}
    I_{i,j,k+1} - I_{i,j,k} &\mbox{if } M_{i,j,k} = M_{i,j,k+1} = 1
\\
    0 & \mbox{otherwise}
\end{cases}
\end{eqnarray*}

The divergence operator for a gradient
$p$ is defined by:
\begin{eqnarray*}
(\mbox{div}~p)^{i,j,k} =
\begin{cases}
    p^x_{i,j,k} - p^x_{i-1,j,k} & \mbox{if } M_{i,j,k} = M_{i-1,j,k} = 1
\\
    p^x_{i,j,k} & \mbox{if } M_{i,j,k} \neq M_{i-1,j,k}= 0
 \\
    -p^x_{i-1,j,k} & \mbox{if } M_{i,j,k} \neq M_{i-1,j,k}=1 \\
\end{cases}
\\
+
\begin{cases}
    p^y_{i,j,k} - p^y_{i,j-1,k} & \mbox{if } M_{i,j,k} = M_{i,j-1,k} = 1
\\
    p^y_{i,j,k} & \mbox{if } M_{i,j,k} \neq M_{i,j-1,k}= 0
 \\
    -p^y_{i,j-1,k} & \mbox{if } M_{i,j,k} \neq M_{i,j-1,k}=1 \\
\end{cases}
\\
+
\begin{cases}
    p^z_{i,j,k} - p^z_{i,j,k-1} & \mbox{if } M_{i,j,k} = M_{i,j,k-1} = 1
\\
    p^z_{i,j,k} & \mbox{if } M_{i,j,k} \neq  M_{i,j,k-1}= 0
 \\
    -p^z_{i,j,k-1} & \mbox{if } M_{i,j,k} \neq M_{i,j,k-1}=1 \\
\end{cases}
\end{eqnarray*}

\section{\emph{ISTA} procedure}
\label{appendix_prox}

We give the sketch of proof of \eqref{Eq:prox_ista}.
The loss $\mathcal{L}(\bold{w})$ being differentiable, the second-order
linearization
of
$\mathcal{L}(\bold{w})$ reads:
\begin{eqnarray*}
\mathcal{L}(\bold{w}) &\approx& \mathcal{L}(\bold{w}^{(k)}) +
(\bold{w}-\bold{w}^{(k)})^T
\nabla
\mathcal{L}(\bold{w}^{(k)})\\
 &+& \frac{1}{2} (\bold{w}-\bold{w}^{(k)})^T\nabla^2 
\mathcal{L}(\bold{w}^{(k)})(\bold{w}-\bold{w}^{(k)})
\end{eqnarray*}
With $L_0$ the Lipschitz constant of $ \nabla
\mathcal{L}$, we have:
$$
\| \nabla  \mathcal{L}(\bold{w}) - \nabla  \mathcal{L}(\bold{w}^{(k)}) \|^2 \leq
L_0
\|\bold{w}-\bold{w}^{(k)}\|^2
$$
Using \cite{ortega2000}, we obtain:
\begin{eqnarray*}
\bold{w}^{(k+1)} =  \mathrm{arg}\!\min_{\bold{w}} \; 
\mathcal{L}(\bold{w}^{(k)})
+ \frac{L}{2}\|\bold{w}-\bold{w}^{(k)}\|^2 + \lambda J(\bold{w})\\
+ (\bold{w}-\bold{w}^{(k)})^T \nabla  \mathcal{L}(\bold{w}^{(k)})
\end{eqnarray*}
Ignoring constant terms, this can be rewritten as:
\begin{eqnarray*}
\bold{w}^{(k+1)} =  \mathrm{arg}\!\min_{\bold{w}} \; \frac{1}{2} \| \bold{w} -
(\bold{w}^{(k)} - \frac{1}{L}
\nabla  \mathcal{L}(\bold{w}^{(k)}) \|^2 + \frac{1}{L} \lambda J(\bold{w}),
\end{eqnarray*}
where $L\geq L_0$ \cite{daubechies2004}.
Finally, using definition \eqref{Def:proximity} of the proximity operator
for $J(\bold{w})$, this is equivalent to \eqref{Eq:prox_ista}:
\begin{equation*}
\bold{w}^{(k+1)} =  \mbox{prox}_{\lambda J /L} \left(\bold{w}^{(k)}
- \frac{1}{L}\nabla  \mathcal{L}(\bold{w}^{(k)}) \right)
\end{equation*}
\section{Dual problem and duality gap computation}
\label{appendix_dualgap}

We give the sketch of proofs of propositions \ref{prop:dual_prox_tv}
and \ref{Prop:dualgap}.
We recall \cite{boyd2004} that the duality between the $\ell_1$ norm and
the $\ell_{\infty}$ norm yields:
\begin{equation}
TV (\bold{v})  = \| \nabla \bold{v} \|_1 = \max_{\|\bold{z}\|_{\infty} \leq 1}
\lb \nabla \bold{v}, \bold{z} \rb
\label{Eq:dualitytv}
\end{equation}
and that the adjoint relation between the gradient and the divergence
operator reads:
\begin{equation}
    \lb \nabla \bold{v}, \bold{z} \rb = - \lb \bold{v}, \dive \, \bold{z} \rb
\label{Eq:adjoint}
\end{equation}

Using \eqref{Eq:dualitytv} and \eqref{Eq:adjoint}, we minimize:
\begin{eqnarray*}
          & & \min_{\bold{v}} \left( \frac{1}{2}\fro{\bold{w}-\bold{v}}^2 +
\lambda TV (\bold{v}) \right)\\
       &=& \lambda \min_{\bold{v}} \left(\frac{1}{2
\lambda}\fro{\bold{w}-\bold{v}}^2 +
\max_{\|\bold{z}\|_{\infty} \leq 1} \lb \nabla \bold{v}, \bold{z} \rb \right)\\
        &=& \lambda \max_{\|\bold{z}\|_{\infty} \leq 1} \left(\min_{\bold{v}}
        \left(\frac{1}{2 \lambda}\fro{\bold{w}-\bold{v}}^2 + \lb \nabla
\bold{v}, \bold{z} \rb
            \right)\right)\\
        &=& \lambda \max_{\|\bold{z}\|_{\infty} \leq 1} \left(\min_{\bold{v}}
        \left(\frac{1}{2 \lambda}\fro{\bold{w}-\bold{v}}^2 - \lb \bold{v},
\dive \, \bold{z} \rb
            \right)\right)\\
\end{eqnarray*}
The computation of the minimum and the maximum above can be exchanged because
the optimization over $\bold{v}$ is convex and the optimization over $\bold{z}$
is concave
\cite{boyd2004}.
\newline
%
By setting the derivative with respect to $\bold{v}$ to $0$ one gets the
resulting
solution of the minimization problem over $\bold{v}$:
\begin{equation*}
\min_{\bold{v}} \left(\frac{1}{2 \lambda}\fro{\bold{w}-\bold{v}}^2 - \lb
\bold{v}, \dive \, \bold{z} \rb
            \right)
\; \Rightarrow \;
    \bold{v}^* = \bold{w} + \lambda \dive \, \bold{z}
\end{equation*}

Replacing $\bold{v}$ by $\bold{v}^*$ in the previous expression leads to:
\begin{eqnarray*}
           & &\min_{\bold{v}} \left( \frac{1}{2}\fro{\bold{w}-\bold{v}}^2 +
\lambda TV (\bold{v}) \right)\\
         &=& \lambda \max_{\|\bold{z}\|_{\infty} \leq 1} \left(
\frac{\lambda}{2}\fro{\dive \, \bold{z}}^2 - \lb \bold{w}, \dive \, \bold{z}\rb
- \lambda
\fro{\dive \, \bold{z}}^2 \right)\\
         &=& \lambda \max_{\|\bold{z}\|_{\infty} \leq 1} \left( -
\frac{\lambda}{2}\fro{\dive \, \bold{z}}^2 - \lb \bold{w}, \dive \, \bold{z} \rb
\right)\\
         &=& \frac{1}{2} \max_{\|\bold{z}\|_{\infty} \leq 1} \left(- \lambda^2
\fro{\dive \, \bold{z}}^2 - 2 \lambda \lb \bold{w}, \dive \, \bold{z} \rb
\right) \\
         &=& \frac{1}{2} \max_{\|\bold{z}\|_{\infty} \leq 1} \left(
\fro{\bold{w}}^2 - \fro{
\lambda \dive \, \bold{z} + \bold{w} }^2 \right)
\end{eqnarray*}
\noindent This gives the proof of Prop.~\ref{prop:dual_prox_tv}.
Also, given a variable $\bold{z}$ satisfying $\|\bold{z}\|_{\infty} \leq 1$ and
an
associated $\bold{w}$ such that $\bold{v} = \bold{w} + \lambda \dive \, \bold{z}
$,
one can guarantee that
\begin{equation*}
    \frac{1}{2}\fro{\bold{w}-\bold{v}}^2 + \lambda TV (\bold{v}) \geq
\frac{1}{2}(\fro{\bold{w}}^2 - \fro{\bold{v}}^2)
\end{equation*}
The strict convexity of the problem guarantees that, at the optimum,
the equality holds.
This last derivation proves the proposition~\ref{Prop:dualgap}.

\end{document}